\documentclass[journal]{IEEEtran}

\ifCLASSINFOpdf
\else
   \usepackage[dvips]{graphicx}
\fi
\usepackage{url}

\usepackage{graphicx}
\usepackage{color}
\usepackage{amsmath}
\usepackage{bm}
\usepackage{float}
\usepackage{booktabs}
\usepackage{cite}
\usepackage{makecell} % for drawing tables
\usepackage{amssymb}
\usepackage{caption}
\usepackage{algorithm}
\usepackage{algorithmic}
\usepackage{multirow}
\usepackage{pifont}
\usepackage{hyperref}
\begin{document}

\title{Tensor-based Graph Learning with Consistency and Specificity for Multi-view Clustering}    

\author{Long Shi, \IEEEmembership{Member, IEEE}, Yunshan Ye, Lei Cao, Yu Zhao, Badong Chen, \IEEEmembership{Senior Member, IEEE}
\thanks{Long Shi, Lei Cao and Yu Zhao are with the School of Computing and Artificial Intelligence, Southwestern University of Finance and Economics, Chengdu 611130, China. They are also affiliated with the Financial Intelligence and Financial Engineering Key Laboratory of Sichuan Province and the Engineering Research Center of Intelligent Finance, Ministry of Education. (e-mail: shilong@swufe.edu.cn, caolei2000@smail.swufe.edu.cn, zhaoyu@swufe.edu.cn)} 
\thanks{Yunshan Ye is with the School of Business Administration, Southwestern University of Finance and Economics, Chengdu 611130, China (e-mail: 1241201z5006@smail.swufe.edu.cn)}
%\thanks{Changqing Zhang is with the College of Intelligence and Computing,
%Tianjin University, Tianjin 300350, China (e-mail: zhangchangqing@tju.edu.cn.)}
\thanks{Badong Chen is with the Institute of Artificial Intelligence and Robotics, Xi’an Jiaotong University, Xi’an 710049, China (e-mail: chenbd@mail.xjtu.edu.cn)}

\thanks{The work of Long Shi was partially supported by the National Natural Science Foundation of China under Grant 62201475 and Sichuan Science and Technology Program under Grant 2024NSFSC1436. The work of Yu Zhao was supported by the National Natural Science Foundation of China under Grant Nos. 62376227, and Sichuan Science and Technology Program under Grant No. 2023NSFSC0032. The work of Badong Chen was supported by the National Natural Science Foundation of China under Grants U21A20485 and 61976175.}

\thanks{This paper has supplementary downloadable material available at http://ieeexplore.ieee.org., provided by the author. The material includes additional experiments to validate the model's performance. This material is 647KB in size.}} 
%The work of Badong Chen was supported by the National Natural Science Foundation of China under Grants U21A20485 and 61976175.}}

\markboth{Journal of \LaTeX\ Class Files}
{Shell \MakeLowercase{\textit{et al.}}: Bare Demo of IEEEtran.cls for IEEE Journals}
\maketitle

\begin{abstract}
In the context of multi-view clustering, graph learning is recognized as a crucial technique, which generally involves constructing an adaptive neighbor graph based on probabilistic neighbors, and then learning a consensus graph for clustering. However, it is worth noting that these graph learning methods encounter two significant limitations. Firstly, they often rely on Euclidean distance to measure similarity when constructing the adaptive neighbor graph, which proves inadequate in capturing the intrinsic structure among data points in practice, particularly for high-dimensional data. Secondly, most of these methods focus solely on consensus graph, ignoring unique information from each view. Although a few graph-based studies have considered using specific information as well, the modelling approach employed does not exclude the noise impact from the common or specific components. To this end, we propose a novel tensor-based multi-view graph learning framework that simultaneously considers consistency and specificity, while effectively eliminating the influence of noise. Specifically, we calculate similarity using pseudo-Stiefel manifold distance to preserve the intrinsic properties of data. By making an assumption that the learned neighbor graph of each view comprises a consistent part, a specific part, and a noise part, we formulate a new tensor-based target graph learning paradigm for noise-free graph fusion. Owing to the benefits of tensor singular value decomposition (t-SVD) in uncovering high-order correlations, this model is capable of achieving a comprehensive understanding of the target graph. Furthermore, we derive an algorithm to address the optimization problem. Experiments on six datasets have demonstrated the superiority of our method. We have released the source code on \href{https://github.com/lshi91/CSTGL-Code}{https://github.com/lshi91/CSTGL-Code}.    
\end{abstract}


%\IEEEpeerreviewmaketitle

\section{Introduction}

\IEEEPARstart{T}{he} advent of the big data era has driven the prominence of multimedia data that originates from diverse sources \cite{zhang2019feature,zhao2017multi}. For example,  multimedia retrieval involves the integral utilization of color, textures, and edges \cite{yang2019skeletonnet}, while video summarization requires the incorporation of multiple videos from distinct perspectives \cite{fu2010multi}. In the context of multimedia retrieval, each type of media data (color, textures, edges) can be considered as a different view. Consequently, multimedia data can be treated as a representative class of multi-view data, which has found extensive applications in multi-view clustering \cite{zhang2022learning, xie2018hyper}.

As a crucial unsupervised task, multi-view clustering overcomes the limitations of single-view clustering, particularly when the view is incomplete or the view data is noisy \cite{chen2021low, shen2023robust}. In contrast to single-view clustering, a particular focus in multi-view clustering is on how to fuse data from multiple views to attain an enhanced data representation \cite{gao2015multi, li2018survey}. Multi-view Subspace Clustering (MVSC) aims to achieve a common subspace representation from multiple views, and then cluster a collection of data points based on the resulting low-dimensional subspaces \cite{liu2012robust,elhamifar2013sparse,peng2016constructing, cao2023robust}. Instead of directly learning the joint representation from original features, latent multi-view methods first extract the latent representation from multi-view data, and then learn the subspace self-representation matrix \cite{zhang2017latent}. These methods effectively address the issue of suboptimal representation in multi-view subspace clustering when data is insufficient \cite{zhang2018generalized, shi2024enhanced}. In addition, tensor-based methods have promoted the advancement of multi-view subspace clustering towards the extraction of high-order data information \cite{xie2018unifying, zhang2020tensorized, gao2020tensor}.

%Further progress in multi-view subspace clustering methods can be found in the preservation of structural information \cite{zhu2019structured, sun2021scalable, liu2023preserving}, robustness inspired by information-theoretic principle \cite{xing2019correntropy,long2023multi,tian2023variational}, and mining of nonlinear data properties \cite{xie2020robust, xie2020joint, shi2024nonlinear}.   

%The fundamental concept behind subspace clustering is the so-called \emph{self-expressiveness} property of the data, which states that each data point within a union of subspaces can be efficiently represented as a linear or affine combination of other points

In addition, graph-based methods also occupy an important position in multi-view clustering. Basically, they conform to the paradigm of constructing neighbor graphs for each view from the original data points \cite{nie2016parameter,nie2017self}, followed by the fusion of these graphs \cite{li2021consensus, liang2019consistency}. Most graph-based methods can be considered as drawing inspiration from \cite{nie2014clustering}. In essence, these methods are variations of learning a consensus graph by fusing adaptive neighbor graphs from different views. Despite a few graph-based studies, such as \cite{liang2019consistency}, have attempted to consider both consistency and specificity, they model the similarity matrix with a consistent part and a specific part, without any further consideration. Specifically, they overlook the impact of noise, which could result in the common and specific information learned being contaminated with noise. To be more clear, these methods have the following limitations:

\emph{1)} The adaptive neighbor graph is constructed using Euclidean distance, which probably fails to characterize the inherent properties of data, especially for high-dimensional data that is commonly encountered in real world scenarios.

\emph{2)} Most methods do not consider view-specific information when learning the consensus graph. Although a few works have made preliminary explorations in this direction, the modeling approach does not exclude the influence of noise from both the common and specific components. For more details, readers are encouraged to refer to the motivation subsection in Section III. 

To address the drawbacks previously discussed, we propose a novel tensor-based graph learning framework, namely Tensor-based Graph Learning with Consistency and Specificity (CSTGL). We begin with learning the neighbor graphs of different views using the pseudo-Stiefel manifold distance. In contrast to Euclidean distance, the newly adopted distance measure is a simplified version of the true Stiefel manifold distance \cite{chakraborty2019statistics}, which is capable of providing a more accurate reflection of the underlying data structure. To facilitate the subsequent graph fusion, we make an assumption that the neighbor graph of each view can be explicitly modelled by a consistent component, a specific component, and an error component. Then, we formulate a new tensor graph learning framework that takes full advantage of both consistent and specific information, while effectively mitigating the influence of noise. Our model can exploit the high-order correlations among graphs, thanks to tensor singular value decomposition (t-SVD) \cite{kilmer2013third}. CSTGL can be treated as a noise-free graph fusion framework, and our main contributions include:

\begin{itemize}
\item We learn the neighbor graph for each view by using the pseudo-Stiefel manifold distance, which allows for a more effective exploitation of the underlying structure compared to Euclidean distance.

\item By assuming that the neighbor graph of each view is composed of a consistent component, a specific component, and an error component, we construct a complete tensor graph fusion framework that explicitly incorporates both consistent and specific tensor graphs, while effectively minimizing the impact of noise.      

\item An iterative algorithm is derived for solving the optimization problem in CSTGL.  
\end{itemize}  

\noindent Experiments demonstrate that CSTGL outperforms some SOTA baselines. Moreover, ablation studies confirm the benefits of using the pseudo-Stiefel manifold distance for similarity measurement, as well as the improvements gained from the integration of consistent and specific tensor graphs.   

The remainder of this paper is organized as follows. Section 2 presents notations and preliminaries. Section 3 shows the detailed procedures of CSTGL. Section 4 carries out extensive experiments to validate CSTGL's performance. We finally draw some conclusions in Section 5.     

\section{Notations and Preliminaries}

In the following, we list the notations used in this paper and introduce some fundamental definitions related to tensor. Moreover, we briefly review some relevant work on subspace-based and graph-based methods.  

\subsection{Notations}

In this paper, we use bold fonts to denote vectors or matrices, and utilize calligraphic fonts to represent tensors. For convenience, we summarize the definitions of notations in Table \ref{table_notation}. For a 3-order tensor $\mathcal{A}\in \mathbb{R}^{n_1\times n_2\times n_3}$, $\mathcal{A}(i,:,:)$ denotes the $i$-th horizontal slice of $\mathcal{A}$, $\mathcal{A}(:,i,:)$ represents the $i$-th lateral slice of $\mathcal{A}$, and $\mathcal{A}(:,:,i)$ accounts for the $i$-th frontal slice of $\mathcal{A}$. Since the $i$-th frontal slice of $\mathcal{A}$ is frequently used in the paper, we define $\mathbf{A}^{(i)} = \mathcal{A}(:,:,i)$. The symbol $\mathbf{A}_{(i)}$ represents the unfolding matrix along the $i$-th mode of $\mathcal{A}$. $\hat{\mathcal{A}} = \mathrm{fft}(\mathcal{A},[\,],3)$ stands for the fast Fourier transformation (FFT) of $\mathcal{A}$ along the 3rd dimension. $\Phi(\mathbf{A^{(1)}},\cdots,\mathbf{A^{(i)}})$ spans a tensor using $\mathbf{A}^{(1)}, \cdots, \mathbf{A}^{(i)}$. The definitions of various norms can be found in the table below.

\begin{table}[h]
\centering 
\caption{Definitions of notations}
\begin{tabular}{l|l}
\Xhline{1.1pt}
Notations & Description \\ 
\hline
$(\cdot)^T$ & Transpose of a vector, a matrix or a tensor\\
\hline
$\mathrm{tr}(\cdot)$ & Trace of a matrix\\
\hline
$\lVert\cdot\rVert_2$ & $l_2$-norm of a vector\\ 
\hline
$a$, $\mathbf{a}$, $\mathbf{A}$, $\mathcal{A}$  & a scale, a vector, a matrix, a tensor \\ 
\hline
$\mathcal{A}(i,:,:)$ & $i$-th horizontal slice of $\mathcal{A}$\\
\hline
$\mathcal{A}(:,i,:)$ & $i$-th lateral slice of $\mathcal{A}$\\
\hline
$\mathcal{A}(:,:,i)$ & $i$-th frontal slice of $\mathcal{A}$\\
\hline
$\hat{\mathcal{A}} = \mathrm{fft}(\mathcal{A},[\,],3)$ & FFT of $\mathcal{A}$ along the 3rd dimension\\
\hline
$\mathcal{A} = \mathrm{ifft}(\hat{\mathcal{A}},[\,],3)$ & Inverse FFT of $\hat{\mathcal{A}}$ along the 3rd dimension\\
\hline
$\mathbf{A^{(i)}}$ & $\mathbf{A^{(i)}} = \mathcal{A}(:,:,i)$\\
\hline
$\hat{\mathbf{A}}^{(i)}$ & $\hat{\mathbf{A}}^{(i)} = \hat{\mathcal{A}}(:,:,i)$\\ 
\hline
$\mathbf{A}_{(i)}$ & Unfolding matrix along the $i$-th mode of $\mathcal{A}$\\
\hline
$\Phi(\mathbf{A^{(1)}},\cdots,\mathbf{A^{(i)}})$ & Span a tensor $\mathcal{A}$ using $\mathbf{A^{(1)}},\cdots,\mathbf{A^{(i)}}$\\  
\hline   
$\lVert\mathbf{A}\rVert_F $ & $F$-norm of $\mathbf{A}$: $\lVert\mathbf{A}\rVert_F = \sqrt{\sum_{ij}A^2_{ij}}$ \\ 
\hline
$\lVert\mathbf{A}\rVert_*$ & Nuclear norm of $\mathbf{A}$: sum of singular values\\ 
\hline
$\lVert\mathbf{A}\rVert_{2,1}$ & $l_{2,1}$-norm of $\mathbf{A}$: $\lVert\mathbf{A}\rVert_{2,1} = \sum_i\lVert\mathbf{x}_i\rVert_2$\\ 
\hline
$\lVert\mathcal{A}\rVert_F$ & $F$-norm of $\mathcal{A}$: $\lVert\mathcal{A}\rVert_F = \sqrt{\sum_{ijk}|\mathcal{A}_{ijk}|^2}$\\
\hline
$\lVert\mathcal{A}\rVert_{\circledast}$ & $t$-SVD based tensor nuclear norm of $\mathcal{A}$\\
\hline
$\lVert\mathcal{A}\rVert_{2,1}$ & $l_{2,1}$-norm of $\mathcal{A}$: $\lVert\mathcal{A}\rVert_{2,1} = \sum_{i,j}\lVert\mathcal{A}(i,j,:)\rVert_2$\\
\Xhline{1.1pt}
\end{tabular}
\label{table_notation}
\end{table} 

\subsection{Fundamental Concepts}

To facilitate the understanding of tensor-related concepts, we introduce the following definitions \cite{kilmer2013third}:

\emph{Definition 1} {\bf{\emph{(t-Product)}}}: Consider two tensors $\mathcal{A}\in\mathbb{R}^{n_1\times n_2\times n_3}$ and $\mathcal{B}\in\mathbb{R}^{n_2\times n_4\times n_3}$, the t-product $\mathcal{A}*\mathcal{B}$ is given by
\begin{equation}
	\label{eq01}
	\mathcal{A}*\mathcal{B} = \mathrm{fold}(\mathrm{bcirc}(\mathcal{A})\mathrm{bvec}(\mathcal{B})), 
\end{equation}
where $\mathrm{bcirc}(\mathcal{A})$ and $\mathrm{bvec}(\mathcal{B})$ denote the block circulant matrix and block vectorizing operations, respectively, defined by
\begin{equation}
	\label{eq02}
	\mathrm{bcirc}(\mathcal{A}) = 
	\begin{bmatrix}
		\mathcal{A}^{(1)} & \mathcal{A}^{(n_3)} & \cdots & \mathcal{A}^{(2)} \\
		\mathcal{A}^{(2)} & \mathcal{A}^{(1)} & \cdots & \mathcal{A}^{(3)} \\
		\vdots & \ddots & \ddots & \vdots\\
		\mathcal{A}^{(n_3)} & \mathcal{A}^{(n_3-1)} & \cdots & \mathcal{A}^{(1)}	
	\end{bmatrix},	
\end{equation}
and
\begin{equation}
	\label{eq03}
	\mathrm{bvec} = [\mathcal{A}^{(1)};\mathcal{A}^{(2)};\cdots;\mathcal{A}^{(n_3)}].
\end{equation} 

\emph{Definition 2} {\bf{\emph{(f-Diagonal Tensor)}}}: A tensor is called f-diagonal if every frontal slice forms a diagonal matrix.

\emph{Definition 3} {\bf{\emph{(Orthogonal Tensor)}}}: For an orthogonal tensor $\mathcal{Q}\in \mathbb{R}^{n\times n\times n_3}$, it satisfies
\begin{equation}
	\label{eq04}
	\mathcal{Q}^T*\mathcal{Q} = \mathcal{Q}*\mathcal{Q}^T = \mathcal{I},
\end{equation}
where $\mathcal{I}$ represents the identity tensor with its frontal slice being a $n\times n$ identify matrix.  

%\emph{Definition 2} {\bf{\emph{(Identity Tensor)}}}: Given an identity tensor $\mathcal{I}\in\mathbb{R}^{n\times n\times n_3}$, its first frontal slice is a $n\times n$ identity matrix, and all other frontal slices are zero matrices.

\emph{Definition 4} {\bf{\emph{(t-SVD)}}}: For a tensor $\mathcal{A}\in\mathbb{R}^{n_1\times n_2\times n_3}$, one can factorize it by t-SVD as
\begin{equation}
	\label{eq05}
	\mathcal{A} = \mathcal{U}*\mathcal{G}*\mathcal{V}^T,
\end{equation}
where $\mathcal{U}\in \mathbb{R}^{n_1\times n_1\times n_3}$ and $\mathcal{V}\in \mathbb{R}^{n_2\times n_2\times n_3}$ are orthogonal, and $\mathcal{G}\in \mathbb{R}^{n_1\times n_2\times n_3}$ is f-diagonal. 

\emph{Definition 5} {\bf{\emph{(t-SVD Based Tensor Nuclear Norm)}}}: The t-SVD based tensor nuclear norm of $\mathcal{A}$ is defined by
\begin{equation}
	\label{eq06}
	\lVert\mathcal{A}\rVert_{\circledast} = \sum^{n_3}_{i=1}\lVert\hat{\mathcal{A}}^{(i)}\rVert_{*} = \sum^{{\rm{min}}(n_1,n_2)}_{j=1}\sum^{n_3}_{i=1}|\hat{\mathcal{G}}(j,j)|,
\end{equation}
where $\hat{\mathcal{G}}^{(i)}$ is calculated by applying the SVD of frontal slices of $\hat{\mathcal{A}}$, i.e., $\hat{\mathcal{A}}^{(i)}=\hat{\mathcal{U}}^{(i)}\hat{\mathcal{G}}^{(i)}\hat{\mathcal{V}}^{(i)T}$.

\subsection{Related Work}

\subsubsection{Subspace-based Methods}

Brbi{\'c} \emph{et al}. jointly enforced the associated constraints on the affinity matrix to exploit both the low-rank and sparse properties of data \cite{brbic2018multi}. The authors further investigated the $l_0$ quasi-norm-based regularization to address the overpenalized problem caused by $l_1$-norm \cite{brbic2018ell_0}. Motivated by the underlying assumption that multiple views intrinsically originate from one shared latent representation, there has been an increasing interest towards the latent representation multi-view methods \cite{zhang2018generalized,xie2019multiview,shi2024enhanced}. In addition, some studies emphasize both consistent and specific representations to fully leverage the information from multi-view data \cite{luo2018consistent, wu2019essential}. Considering that weakly supervised information can likely be acquired in real-world scenarios, Tang \emph{et al}. incorporated the weakly supervised information when formulating the regularization constraints \cite{tang2021constrained}. Applying kernel methods to subspace clustering enhances the data fusion from multiple views in nonlinear subspaces \cite{xie2020robust, wang2021late}. In tensor-based multi-view methods, imposing low-rank constraints on tensors helps reduce redundancy among different views \cite{wu2020unified, chen2022low, jiang2022tensorial}. This is achieved by capturing essential information and discarding irrelevant or redundant features \cite{jiang2022tensorial, xia2022tensorized}. Some researchers have also explored low-rank tensor approximation to enhance learning effectiveness in the presence of noise \cite{che2024enhanced,pu2023robust,wang2021error}. Further progress in multi-view subspace clustering methods can be found in the preservation of structural information \cite{zhu2019structured, sun2021scalable, liu2023preserving}, robustness inspired by information-theoretic principle \cite{xing2019correntropy,long2023multi,tian2023variational}, and mining of nonlinear data properties \cite{xie2020joint, shi2024nonlinear}.  

\subsubsection{Graph-based Methods} Nie \emph{et al}. explored the assignment of weights to different graphs for highlighting the significance of key views \cite{nie2016parameter, nie2017self}. These methods follow the technical approach of pursuing consistency for the target graph, which has inspired several variants \cite{kang2020multi, ren2020consensus, li2021consensus,  wang2022towards}. Research has shown that establishing a unified graph matrix from different views is useful to integrate comprehensive information \cite{wang2019gmc, liang2019consistency}. In order to reduce dimensionality, a parameter-free method that allows for structured graph learning was investigated in \cite{wang2019parameter}. In \cite{kang2021structured}, Kang \emph{et al}. presented a graph learning framework that is able to preserve a complete structure of data. In addition, some research works have tackled other issues. For example, Wen \emph{et al}. investigated the graph completion issue for incomplete views \cite{wen2020adaptive} to explore hidden information in missing views. Xing \emph{et al}. proposed learning the consensus graph via correntropy to enhance robustness in the presence of missing views. To deal with large-scale datasets, Xia \emph{et al}. focused on bipartite graph learning to extract fundamental graph information \cite{xia2022tensorized}. Wang \emph{et al}. designed a scalable and flexible anchor graph fusion framework to unify the procedures of anchor graph construction and individual graph fusion, as well provide a more reasonable anchor selection scheme by considering data diversity \cite{wang2024scalable}.

\section{Proposed Methodology}

\begin{figure*}[t]
\centering
\includegraphics[width=1\textwidth]{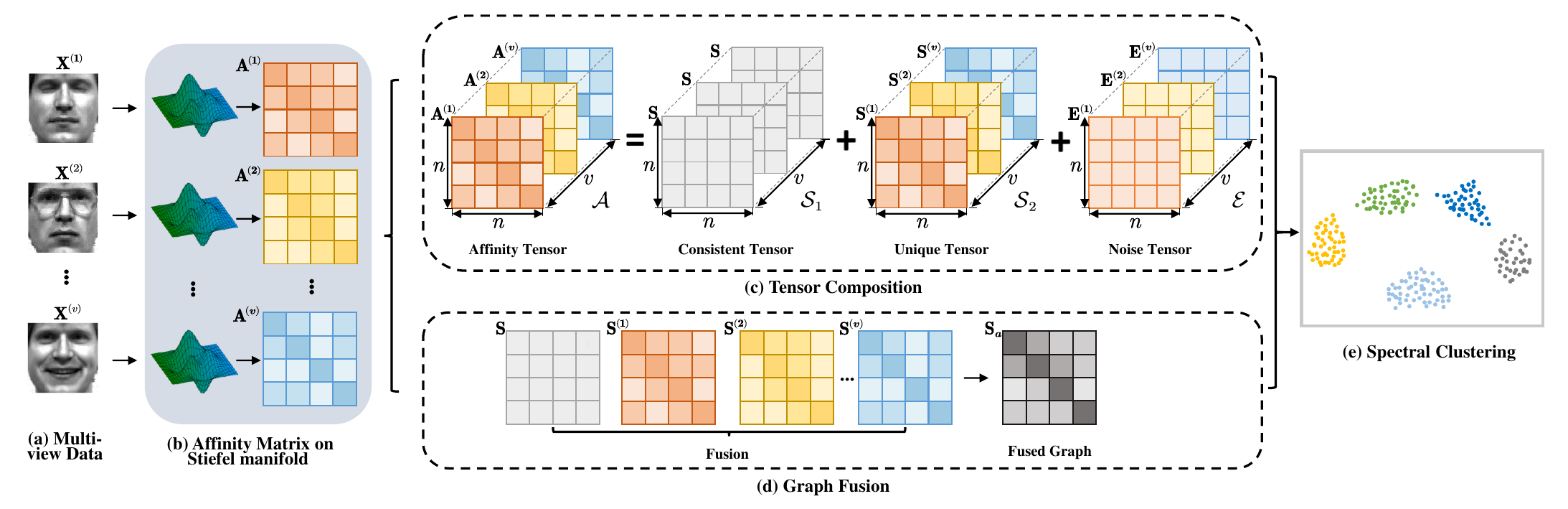} % Reduce the figure size so that it is slightly narrower than the column. Don't use precise values for figure width.This setup will avoid overfull boxes.
\caption{Framework of CSTGL.}
\label{Fig_fra}
\end{figure*}

We will introduce CSTGL, offering a detailed presentation of its motivation, model formulation, optimization procedures and computational complexity analysis. The framework of CSTGL is shown in Fig. \ref{Fig_fra}.

\subsection{Motivation}

Our proposed CSTGL is motivated by the following three observations from graph-based methods:

\emph{Observation 1:} In order to reflect the similarity among data samples, a common strategy is to learn the adaptive neighbor graph for $\mathbf{A}^{(v)}$ \cite{li2021consensus, chen2022low}. However, most methods construct the adaptive neighbor graph by Euclidean distance, which may not effectively uncover the inherent structure of high-dimensional data, since using Euclidean distance inevitably suffers from the curse of dimensionality.   

\emph{Observation 2:} We also notice that the majority of graph-based multi-view methods follow the technical approach of learning a consensus graph for the target graph by minimizing the discrepancy between the similarity graph and the target graph. This learning strategy notably emphasizes the importance of consistency among views. However, from another perspective, it neglects the diversity inherent in different views.  

\emph{Observation 3:} Despite a few graph-based works, such as \cite{liang2019consistency}, have attempted to consider both consistency and specificity, they model the similarity matrix $\mathbf{A}^{(v)}$ with a consistent part $\mathbf{S}$ and a specific part $\mathbf{S}^{(v)}$, i.e., $\mathbf{A}^{(v)} = \mathbf{S} + \mathbf{S}^{(v)}$. However, this model exists a significant limitation. It does not take into account the impact of noise, which probably leads to the noise being assigned to both the consistent and specific components during the learning process. Ideally, either the consistent or specific component should capture useful information rather than containing noise. Therefore, it is necessary to explicitly distinguish noise from useful components when modeling.

In order to tackle these limitations in the above observations, our goal is to design a novel graph learning framework that is equipped with an improved similarity measure for $\mathbf{A}^{(v)}$, as well as a noise-free graph fusion model that considers both consistency and specificity.    

\subsection{Model Formulation}

For the purpose of capturing the intrinsic similarity between data points, we learn the neighbor graph for each view by measuring the similarity distance on the Stiefel manifold. Before proceeding, we briefly introduce the distance measure on the Stiefel manifold. 

\emph{Stiefel manifold:} The Stiefel manifold represents the set of matrices with orthonormal columns. The Stiefel manifold is commonly used in optimization and geometry, particularly in problems involving rotations and orthogonal transformations. Given two $d\times n$ orthonormal matrices $\mathbf{X}$ and $\mathbf{Y}$, the $l_1$-norm geodesic distance between them is defined as \cite{chen2020estimating}
\begin{equation}
	\label{eq09}
	d_s(\mathbf{X},\mathbf{Y}) = d - \mathrm{tr}(\mathbf{X}\mathbf{Y}^T).
\end{equation}

The above equation characterizes the distance between two matrices $\mathbf{X}$ and $\mathbf{Y}$. However, our goal is to measure the distance between two samples, i.e., two vectors, rather than between matrices. Consequently, it is necessary to adapt Eq. (\ref{eq09}) to align with our requirements. A straightforward approach is to set $n=1$, reducing the distance calculation in Eq. (\ref{eq09}) to a form suitable for vectors. The resulting expression is as follows:
%Inspired from (\ref{eq09}), a natural variant that involves two vectors $\mathbf{x}$ and $\mathbf{y}$ can be formulated by
\begin{equation}
	\label{eq10}
	d_s(\mathbf{x},\mathbf{y}) = d - \mathrm{tr}(\mathbf{x}\mathbf{y}^T),
\end{equation}
where $\mathbf{x}$ and $\mathbf{y}$ are two $d\times 1$ vectors. Strictly speaking, since $\mathbf{x}$ and $\mathbf{y}$ are not necessarily unit vectors, this formulation does not rigorously adhere to the definition of the geodesic distance on the Stiefel manifold for the case of $n=1$. Although our distance measure is not a true Stiefel manifold distance, it draws inspiration from Stiefel manifold distance. Therefore, we refer to the distance measure in Eq. (\ref{eq10}) as the ``pseudo-Stiefel manifold distance".

\textbf{\textit{Remark 1:}} From (\ref{eq10}), we have the subsequent observations. If $\mathrm{tr}(\mathbf{x}\mathbf{y}^T) = 0$, then $\mathbf{x}$ and $\mathbf{y}$ are orthonormal to each other, implying that there is no similarity between them. If $\mathrm{tr}(\mathbf{x}\mathbf{y}^T)$ is not zero, then $\mathbf{x}$ and $\mathbf{y}$ are not orthonormal to each other, and there exists some similarity between them. In summary,  (\ref{eq10}) can reflect the similarity between $\mathbf{x}$ and $\mathbf{y}$. The greater the deviation between $d$ and $\mathrm{tr}(\mathbf{x}\mathbf{y}^T)$, the lower the similarity between $\mathbf{x}$ and $\mathbf{y}$. Instead, smaller deviation implies higher similarity. 

Given a set of multi-view data matrices $\{\mathbf{X}^{(v)}\}^m_{v=1}$, with $\mathbf{X}^{(v)}\in \mathbb{R}^{d_v\times n}$ denoting the $v$-th view data with $d_v$ dimensions, we can define a new formulation to learn the neighbor graph $\mathbf{A}^{(v)}$, that is
\begin{equation}
\begin{aligned}
	\label{eq11}
	&\underset{\mathbf{A}^{(v)}}{\mathrm{min}}\sum^m_{v=1}\sum^n_{i,j=1}(d_v-\mathrm{tr}(\mathbf{x}^{(v)}_i(\mathbf{x}^{(v)}_j)^T))a^2_{ij} + \alpha\lVert\mathbf A^{(v)}\rVert^2_F,\\
	& s.t. \; 0\leq a^{(v)}_{ij}\leq 1, (\mathbf{a}^{(v)}_i)^{T}\mathbf{1} = 1,	
\end{aligned}
\end{equation}
where $\mathbf{x}^{(v)}_i$ represents the $i$-th sample of $\mathbf{X}^{(v)}$, $\alpha$ denotes the positive balance parameter, $\mathbf{A}^{(v)}\in \mathbb{R}^{n\times n}$ stands for the similarity matrix, and $\mathbf{a}^{(v)}_i\in \mathbb{R}^{n\times 1}$ is a column vector with its $j$-th element being $a^{(v)}_{ij}$. The main difference between the traditional neighbor graph learning and our method exists in the distance measure. In contrast to the traditional approach that utilizes Euclidean distance $\lVert\mathbf{x}^{(v)}_i - \mathbf{x}^{(v)}_j\rVert^2_2$ to measure similarity, our method uses pseudo-Stiefel manifold distance, which more precisely reflects the intrinsic structure for high-dimensional data. In the supplementary material, we present a toy experiment that demonstrates the superiority of using the pseudo-Stiefel manifold distance for high-dimensional data compared to the Euclidean distance.

After learning the neighbor graphs of all the views, we need to fuse these neighbor graphs. As mentioned in \emph{Observations 2} and \emph{3} of the motivation subsection, most graph-based multi-view methods directly learn a consensus graph, ignoring the specific information of each view. While a few works consider both consistency and specificity, they fail to distinguish noise from useful information. To address these limitations, we suppose that each neighbor graph $\mathbf{A}^{(v)}$ can be modelled by a consistent component, a view-specific component, and a noise component, i.e.,
\begin{equation}
	\label{eq12}
	\mathbf{A}^{(v)} = \mathbf{S} + \mathbf{S}^{(v)} + \mathbf{E}^{(v)}, 
\end{equation} 
where $\mathbf{S}$ denotes the consistent part, $\mathbf{S}^{(v)}$ represents the view-specific component, and $\mathbf{E}^{(v)}$ is the noise term. 

For capturing the high-order correlation among different neighbor graphs \cite{chen2019jointly}, we formulate the following tensor-based graph fusion framework:
\begin{equation}
\begin{aligned}
	\label{eq13}
	&\underset{\mathbf{A}^{(v)},\mathcal{E},\mathcal{S}_1,\mathcal{S}_2}{\mathrm{min}}\sum^m_{v=1}\sum^n_{i,j=1}(d_v-\mathrm{tr}(\mathbf{x}^{(v)}_i(\mathbf{x}^{(v)}_j)^T)) a^2_{ij} + \alpha\lVert\mathbf A^{(v)}\rVert^2_F\\
	&\qquad\qquad + \lVert\mathcal E\rVert_{2,1} + \beta\lVert\mathcal S_1\rVert_{\circledast} + \gamma \lVert\mathcal{S}_2\rVert^2_F\\
	& s.t. \; 0\leq a^{(v)}_{ij}\leq 1, (\mathbf{a}^{(v)}_i)^{T}\mathbf{1} = 1, \mathbf{A}^{(v)} = \mathbf{S} + \mathbf{S}^{(v)} + \mathbf{E}^{(v)}, 
	\end{aligned}
\end{equation} 
where $\mathcal{E} = \Phi(\mathbf{E}^{(1)}, \cdots, \mathbf{E}^{(v)})$ is the noise tensor, $\mathcal{S}_1 = \Phi(\mathbf{S}, \cdots, \mathbf{S})$ denotes the consistent tensor, $\mathcal{S}_2 = \Phi(\mathbf{S}^{(1)}, \cdots, \mathbf{S}^{(v)})$ represents the specific tensor, they are all of size $n\times n\times m$, $\beta$ and $\gamma$ are trade-off parameters. 

\textbf{\textit{Remark 2:}} We impose $l_{2,1}$-norm on the noise tensor $\mathcal{E}$ to encourage the columns of each frontal slice $\mathbf{E}^{(v)}$ to be zero, which is expected to minimize the impact of noise \cite{liu2012robust}. We enforce the consistent tensor $\mathcal{S}_1$ with the $t$-SVD based tensor nuclear norm to exploit more shared high-order information across various neighbor graphs \cite{xie2018unifying}. Moreover, we impose $F$-norm on the specific tensor $\mathcal{S}_2$ in order to ensure the preservation of view-specific information. 

\textbf{\textit{Remark 3:}} In contrast to the model presented in \emph{Observation 3}, our model explicitly separates the noise term from the consistent and specific components, which guarantees that the consistent and specific information learned is useful and free from noise. In essence, our modelling technique contributes to noise-free graph fusion. To demonstrate our arguments, we conduct experiments by plotting the probability density function curves for the learned noise-free information and error term under two situations: in the absence of synthetic noise and in the presence of synthetic noise. Additionally, to further validate that noise information is effectively captured in the error component $\mathbf{E}^{(v)}$, we compare two cases for the fused graph matrix: one incorporating only useful information and the other including the error component. The experimental results and their corresponding analysis are provided in the supplemental material.

\subsection{Optimization}

To handle the optimization problem in (\ref{eq13}), we utilize the Alternating Direction Method of Multipliers (ADMM) \cite{boyd2011distributed} that is commonly used for solving multivariate constrained optimization problems. The core idea of ADMM is to solve each variable-induced subproblem by holding the other variables constant. 

To proceed, we need to introduce two auxiliary variables to make our objective function separable. Let $\mathcal{E} = \mathcal{W}$ and $\mathcal{S}_1 = \mathcal{K}$, the original objective function can be equivalently described by
\begin{equation}
\begin{aligned}
	\label{eq014}
	&\underset{\mathbf{A}^{(v)},\mathcal{W},\mathcal{K},\mathcal{S}_2, \mathcal{S}_1,\mathcal{E}}{\mathrm{min}}\sum^m_{v=1}\sum^n_{i,j=1}(d_v-\mathrm{tr}(\mathbf{x}^{(v)}_i(\mathbf{x}^{(v)}_j)^T)) a^2_{ij}\\
	&\qquad\qquad + \alpha\lVert\mathbf A^{(v)}\rVert^2_F + \lVert\mathcal W\rVert_{2,1} + \beta\lVert\mathcal K\rVert_{\circledast} + \gamma \lVert\mathcal{S}_2\rVert^2_F\\
	& s.t. \; 0\leq a^{(v)}_{ij}\leq 1, (\mathbf{a}^{(v)}_i)^{T}\mathbf{1} = 1, \mathbf{A}^{(v)} = \mathbf{S} + \mathbf{S}^{(v)} + \mathbf{E}^{(v)},\\
	& \qquad \mathcal{E} = \mathcal{W}, \mathcal{S}_1 = \mathcal{K}. 	 
\end{aligned}	
\end{equation}     

\noindent As (\ref{eq12}) can also be written as a tensor-shaped model, i.e., $\mathcal{A} = \mathcal{S}_1 + \mathcal{S}_2 + \mathcal{E}$, the corresponding augmented Lagrangian function of (\ref{eq014}) is given by
\begin{equation}
\begin{aligned}
	\label{eq15}
	&\underset{\mathbf{A}^{(v)},\mathcal{W},\mathcal{K},\mathcal{S}_2, \mathcal{S}_1,\mathcal{E}}{\mathrm{min}}\sum^m_{v=1}\sum^n_{i,j=1}(d_v-\mathrm{tr}(\mathbf{x}^{(v)}_i(\mathbf{x}^{(v)}_j)^T)) a^2_{ij}\\
	&\qquad\qquad + \alpha\lVert\mathbf A^{(v)}\rVert^2_F + \lVert\mathcal W\rVert_{2,1} + \beta\lVert\mathcal K\rVert_{\circledast} + \gamma \lVert\mathcal{S}_2\rVert^2_F\\
	&\qquad\qquad+ \frac{\rho}{2}\lVert \mathcal{S}_1 + \mathcal{S}_2 + \mathcal{E} - (\mathcal{A} + \frac{\mathcal{Q}_1}{\rho})\rVert^2_F\\
	&\qquad\qquad+ \frac{\rho}{2}\lVert\mathcal{K} - (\mathcal{S}_1+\frac{\mathcal{Q}_2}{\rho})\rVert^2_F\\
	&\qquad\qquad+ \frac{\rho}{2}\lVert \mathcal{W} - (\mathcal{E}+\frac{\mathcal{Q}_3}{\rho})\rVert^2_F\\	
	& s.t. \; 0\leq a^{(v)}_{ij}\leq 1, (\mathbf{a}^{(v)}_i)^{T}\mathbf{1} = 1,
\end{aligned}	
\end{equation}
where $\mathcal{Q}_1$, $\mathcal{Q}_2$, $\mathcal{Q}_3$ are the Lagrange multipliers with size $n\times n\times m$, and $\rho$ is a positive factor. By applying the alternative minimization strategy, we decompose the optimization problem in (\ref{eq15}) into the subsequent subproblems.

1) $\mathbf{A}^{(v)}-$\textbf{subproblem} 

We treat $\mathbf{A}^{(v)}$ as a variable while fixing other variables, resulting in
\begin{equation}
\begin{aligned}
	\label{eq016}
	&\underset{0\leq a^{(v)}_{ij}\leq 1, (\mathbf{a}^{(v)}_i)^{T}\mathbf{1} = 1}{\mathrm{min}}\sum^m_{v=1}\sum^n_{i,j=1}(d_v-\mathrm{tr}(\mathbf{x}^{(v)}_i(\mathbf{x}^{(v)}_j)^T)) a^2_{ij}\\
	&\qquad\qquad + \alpha\lVert\mathbf A^{(v)}\rVert^2_F + \frac{\rho}{2}\lVert \mathcal{S}_1 + \mathcal{S}_2 + \mathcal{E} - (\mathcal{A} + \frac{\mathcal{Q}_1}{\rho})\rVert^2_F
\end{aligned}
\end{equation}  

\noindent By denoting $\mathbf{B}^{(v)} = \mathbf{S} + \mathbf{S}^{(v)} + \mathbf{E}^{(v)} - \frac{\mathbf{Q}^{(v)}_1}{\rho}$ and $\mathbf{b}^{(v)}_i$ as a vector with its $j$-th entry being $b^{(v)}_{ij}$, the above equation can be written as
\begin{equation}
\begin{aligned}
	\label{eq017}
	&\underset{0\leq a^{(v)}_{ij}\leq 1, (\mathbf{a}^{(v)}_i)^{T}\mathbf{1} = 1}{\mathrm{min}}\sum^n_{i,j=1}(d_v-\mathrm{tr}(\mathbf{x}^{(v)}_i(\mathbf{x}^{(v)}_j)^T)) a^2_{ij}\\
	&\qquad\qquad\qquad+ \alpha(a^{(v)}_{ij})^2 + \frac{\rho}{2}(a^{(v)}_{ij} - b^{(v)}_{ij})^2.
\end{aligned}
\end{equation}   

Let $e^{(v)}_{ij} = d_v-\mathrm{tr}(\mathbf{x}^{(v)}_i(\mathbf{x}^{(v)}_j))^T$, we can further reformulate the problem in (\ref{eq017}) separately for each $i$  
\begin{equation}
\begin{aligned}
	\label{eq018}
	&\underset{0\leq a^{(v)}_{ij}\leq 1, (\mathbf{a}^{(v)}_i)^{T}\mathbf{1} = 1}{\mathrm{min}} (\mathbf{e}^{(v)}_i)^T \mathbf{a}^{(v)}_i + \alpha\lVert\mathbf{a}^{(v)}_i\rVert^2_2\\
	&\qquad\qquad\qquad + \frac{\rho}{2}(\lVert\mathbf{a}^{(v)}_i\rVert^2_2 - 2(\mathbf{a}^{(v)}_i)^T \mathbf{b}^{(v)}_i + \lVert\mathbf{b}^{(v)}_i\rVert^2_2),
\end{aligned}
\end{equation}
where $\mathbf{e}^{(v)}_i$ is a vector with its $j$-th entry being $e^{(v)}_{ij}$. 

It is easy to derive that the optimization problem in (\ref{eq018}) can be transformed to solve 
\begin{equation}
\begin{aligned}
	\label{eq019}
	\underset{0\leq a^{(v)}_{ij}\leq 1, (\mathbf{a}^{(v)}_i)^{T}\mathbf{1} = 1}{\mathrm{min}}	\left\lVert \mathbf{a}^{(v)}_i - \frac{\rho\mathbf{b}^{(v)}_i - \mathbf{e}^{(v)}_i}{2\alpha + \rho}\right\rVert^2_2	
\end{aligned}
\end{equation}

\noindent Since the problem in (\ref{eq019}) is an Euclidean projection problem on the simplex space, we rewrite it as
\begin{equation}
\begin{aligned}
	\label{eq020}
	\mathcal{L}(\mathbf{a}^{(v)}_i,\eta,\boldsymbol{\psi})=&\frac{1}{2}\left\lVert \mathbf{a}^{(v)}_i - \frac{\rho\mathbf{b}^{(v)}_i - \mathbf{e}^{(v)}_i}{2\alpha + \rho}\right\rVert^2_2\\
	 &\qquad- \eta((\mathbf{a}^{v}_i)^T\mathbf{1} - 1) - \boldsymbol{\psi}^T\mathbf{a}^{(v)}_i,
\end{aligned}
\end{equation}     
where $\eta$ is a positive constant, and $\boldsymbol{\psi}$ represents a Lagrangian coefficient vector. By applying the Karush-Kuhn-Tucker condition \cite{boyd2004convex}, the final solution to (\ref{eq020}) is 
\begin{equation}
	\label{eq021}
	\mathbf{a}^{(v)^*}_i = \left(\frac{\rho\mathbf{b}^{(v)}_i - \mathbf{e}^{(v)}_i}{2\alpha + \rho} + \eta\mathbf{1}\right)_{+},	
\end{equation}
where $(x)_{+} = \max(x,0)$. After obtaining all $\mathbf{a}^{(v)}_i$, we can reshape them into the matrix or tensor form.

2) $\mathcal{W}-$\textbf{subproblem} 

By keeping $\mathcal{W}$ as a variable and maintaining other variables fixed, we arrive at
\begin{equation}
	\label{eq022}
	\underset{\mathcal{W}}{\mathrm{min}}\;\lVert\mathcal{W}\rVert_{2,1} + \frac{\rho}{2}\lVert \mathcal{W} - (\mathcal{E} + \frac{\mathcal{Q}_3}{\rho})\rVert^2_F.
\end{equation} 

\noindent Given that the $l_{2,1}$-norm of a tensor is defined as the sum of the $l_2$-norms for each fiber along the third mode, thus we have $\lVert\mathbf{W}_{(3)}\rVert_{2,1} = \lVert\mathcal{W}\rVert_{2,1}$. Similarly, according to the definition of the $F$-norm of a tensor, it is straightforward to see $\lVert\mathbf{W}_{(3)} - (\mathbf{E}_{(3)} + \frac{\mathbf{Q}_{3_{(3)}}}{\rho})\rVert^2_F = \lVert \mathcal{W} - (\mathcal{E} + \frac{\mathcal{Q}_3}{\rho})\rVert^2_F$. Thus, the problem in (\ref{eq022}) is transformed to
\begin{equation}
	\label{eq023}
	\underset{\mathbf{W}_{(3)}}{\mathrm{min}}\;\lVert\mathbf{W}_{(3)}\rVert_{2,1} + \frac{\rho}{2}\lVert\mathbf{W}_{(3)} - (\mathbf{E}_{(3)} + \frac{\mathbf{Q}_{3_{(3)}}}{\rho})\rVert^2_F.	
\end{equation}

Let $\mathbf{C} = \mathbf{E}_{(3)} + \frac{\mathbf{Q}_{3_{(3)}}}{\rho}$, and according to \cite{liu2012robust}, the problem in (\ref{eq023}) takes the following closed-form solution
\begin{equation}
	\label{eq024}
	\mathbf{W}^{*}_{(3):,i} = 
	\begin{cases}
		\frac{\lVert\mathbf{C}_{:,i}\rVert_2 - \frac{1}{\rho}}{\lVert\mathbf{C}_{:,i}\rVert_2}, & \rm{if} \; \lVert\mathbf{C}_{:,i}\rVert_2>\frac{1}{\rho}\\
		0, & \rm{otherwise},
	\end{cases}	
\end{equation}
where $\mathbf{C}_{:,i}$ denotes the $i$-th column of $\mathbf{C}$. After obtaining $\mathbf{W}^{*}_{(3):,i}$ of all columns, we can reshape them into the tensor form.

3) $\mathcal{K}-$\textbf{subproblem}

By ignoring irrelevant variables, we update $\mathcal{K}$ by solving
\begin{equation}
	\label{eq025}
	\underset{\mathcal{K}}{\mathrm{min}}\; \beta\lVert\mathcal{K}\rVert_{\circledast} + \frac{\rho}{2}\lVert\mathcal{K} - (\mathcal{S}_1+\frac{\mathcal{Q}_2}{\rho})\rVert^2_F.
\end{equation}

\noindent Let $\mathcal{F} = \mathcal{S}_1+\frac{\mathcal{Q}_2}{\rho}$, the problem in (\ref{eq025}) can be solved by applying the tensor tubal-shrinkage operator \cite{hu2016twist}, which takes the following closed-form solution:
\begin{equation}
	\label{eq026}
	\mathcal{K}^{*} = \mathcal{D}_{\frac{m\beta}{\rho}}(\mathcal{F}) = \mathcal{U}*\mathcal{D}_{\frac{m\beta}{\rho}}(\mathcal{G})*\mathcal{V}^{T},
\end{equation}
where $\mathcal{F} = \mathcal{U}*\mathcal{G}*\mathcal{V}^T$, $\mathcal{D}_{\frac{m\beta}{\rho}}(\mathcal{G}) = \mathcal{G}*\mathcal{P}$, and $\mathcal{P}$ is a f-diagonal tensor with its diagonal element in the Fourier domain being $\hat{\mathcal{P}}(i,i,j)= \rm{max}(1-\frac{m\beta/\rho}{\hat{\mathcal{G}}(i,i,j)},0)$.
    
4) $\mathcal{S}_2-$\textbf{subproblem}

In order to update $\mathcal{S}_2$, we neglect irrelevant variables, and arrive at

\begin{equation}
	\label{eq027}
	\underset{\mathcal{S}_2}{\mathrm{min}}\; \gamma \lVert\mathcal{S}_2\rVert^2_F + \frac{\rho}{2}\lVert \mathcal{S}_1 + \mathcal{S}_2 + \mathcal{E} - (\mathcal{A} + \frac{\mathcal{Q}_1}{\rho})\rVert^2_F.	
\end{equation}

\noindent Motivated from \cite{wang2021multi}, the problem in (\ref{eq027}) can be transformed to solve an equivalent problem in the Fourier domain:
\begin{equation}
	\label{eq028}
	\underset{\hat{\mathcal{S}}_2}{\mathrm{min}}\; \gamma \lVert\hat{\mathcal{S}}_2\rVert^2_F + \frac{\rho}{2}\lVert \hat{\mathcal{S}}_2 - (\hat{\mathcal{A}} + \frac{\hat{\mathcal{Q}}_1}{\rho} - \hat{\mathcal{S}}_1 - \hat{\mathcal{E}})\rVert^2_F.	
\end{equation}

To go further, the solution to (\ref{eq028}) can be derived slice-by-slice along the frontal direction
\begin{equation}
	\label{eq029}
	\underset{\hat{\mathbf{S}}^{(v)}_2}{\mathrm{min}}\; \gamma \lVert\hat{\mathbf{S}}^{(v)}_2\rVert^2_F + \frac{\rho}{2}\lVert \hat{\mathbf{S}}^{(v)}_2 - (\hat{\mathbf{A}}^{(v)} + \frac{\hat{\mathbf{Q}}^{(v)}_1}{\rho} - \hat{\mathbf{S}}^{(v)}_1 - \hat{\mathbf{E}}^{(v)})\rVert^2_F.	
\end{equation}  

\noindent By differentiating (\ref{eq029}) and setting the result to zero, we have
\begin{equation}
	\label{eq030}
	\hat{\mathbf{S}}^{(v)}_2 = \frac{\rho(\hat{\mathbf{A}}^{(v)} + \frac{\hat{\mathbf{Q}}^{(v)}_1}{\rho} - \hat{\mathbf{S}}^{(v)}_1 - \hat{\mathbf{E}}^{(v)})}{2\gamma + \rho}.
\end{equation}

\noindent After getting $\hat{\mathbf{S}}^{(v)}_2$, we obtain $\mathbf{S}^{(v)}_2$ by performing inverse FFT. Then, the tensor $\mathcal{S}^{*}_2$ to be updated can be recovered from $\mathbf{S}^{(v)}_2$.

5) $\mathcal{S}_1-$\textbf{subproblem}

By eliminating other variables except for $\mathcal{S}_1$-related variables, we formulate the following optimization problem
\begin{equation}
	\label{eq031}
	\underset{\mathcal{S}_1}{\mathrm{min}}\; \frac{\rho}{2}\lVert \mathcal{S}_1 + \mathcal{S}_2 + \mathcal{E} - (\mathcal{A} + \frac{\mathcal{Q}_1}{\rho})\rVert^2_F + \frac{\rho}{2}\lVert\mathcal{K} - (\mathcal{S}_1+\frac{\mathcal{Q}_2}{\rho})\rVert^2_F.
\end{equation}  

\noindent In analogy to the problem solving in (\ref{eq027}), we transform the original problem into an equivalent problem in the Fourier domain. By solving this equivalent problem, we obtain
\begin{equation}
	\label{eq032}
	\hat{\mathbf{S}}^{(v)}_1 = \frac{\hat{\mathbf{K}}^{(v)} - \frac{\hat{\mathbf{Q}}^{(v)}_2}{\rho}-[\hat{\mathbf{S}}^{(v)}_2+\hat{\mathbf{E}}^{(v)}-(\hat{\mathbf{A}}^{(v)}+\frac{\hat{\mathbf{Q}}^{(v)}_1}{\rho})]}{2}.
\end{equation} 

\noindent In a similar fashion, the tensor $\mathcal{S}^{*}_1$ can be recovered from $\hat{\mathbf{S}}^{(v)}_1$.

6) $\mathcal{E}-$\textbf{subproblem}

To update $\mathcal{E}$, we discard irrelevant variables and formulate an optimization problem as follows:
\begin{equation}
	\label{eq033}
	\underset{\mathcal{E}}{\mathrm{min}}\; \frac{\rho}{2}\lVert \mathcal{S}_1 + \mathcal{S}_2 + \mathcal{E} - (\mathcal{A} + \frac{\mathcal{Q}_1}{\rho})\rVert^2_F + \frac{\rho}{2}\lVert \mathcal{W} - (\mathcal{E}+\frac{\mathcal{Q}_3}{\rho})\rVert^2_F.
\end{equation}
  
\noindent Following the similar procedures of solving subproblems $\mathcal{S}_2$ and $\mathcal{S}_1$, the solution to (\ref{eq033}) is given by
\begin{equation}
	\label{eq034}
	\hat{\mathbf{E}}^{(v)} = \frac{\hat{\mathbf{A}}^{(v)} + \frac{\hat{\mathbf{Q}}^{(v)}_1}{\rho} + \hat{\mathbf{W}}^{(v)} - \hat{\mathbf{S}}^{(v)}_1 - \hat{\mathbf{S}}^{(v)}_2 - \frac{\hat{\mathbf{Q}}^{(v)}_3}{\rho}}{2}.
\end{equation}  
From (\ref{eq034}), we are able to construct the tensor $\mathcal{E}^*$ that needs to be updated.

7) \textbf{Updating multipliers}

By performing the gradient ascent operation, we update the Lagrange multipliers as follows
\begin{equation}
	\label{eq035}
	\left\{
	\begin{aligned}
		&\mathcal{Q}_1 = \mathcal{Q}_1 + \rho(\mathcal{A}-(\mathcal{S}_1+\mathcal{S}_2)-\mathcal{E})\\
		&\mathcal{Q}_2 = \mathcal{Q}_2 + \rho(\mathcal{S}_1-\mathcal{K})\\
		&\mathcal{Q}_3 = \mathcal{Q}_3 + \rho(\mathcal{E}-\mathcal{W})\\				 
	\end{aligned}
	\right.	
\end{equation}
where $\rho = \mu\rho$ with $\mu$ being a positive value.

After obtaining the learned consistent tensor $\mathcal{S}_1$  and view-specific tensor $\mathcal{S}_2$, we require to fuse them to achieve the final graph matrix $\mathbf{S}_a$. A natural fusion scheme is to average the accumulation of all the frontal slices of $\mathcal{S}_1$ and $\mathcal{S}_2$, which is defined by
\begin{equation}
	\label{eq036}
	\mathbf{S}_a = \frac{1}{m}\sum^m_{v=1}(\mathbf{S}^{(v)}_1+\mathbf{S}^{(v)}_2).	
\end{equation}  
After obtaining the learned fusion graph $\mathbf{S}_a$, we construct the affinity matrix $\mathbf{S}_{af}$ using $\mathbf{S}_{af}=\frac{\mathbf{S}_a+\mathbf{S}^T_a}{2}$, and then perform clustering. For the sake of clarity, we summarize the procedures of executing CSTGL in the Algorithm \ref{alg: learning S}. 

\textbf{\textit{Remark 4:}} Since our model is tensor-based, we cannot strictly guarantee that the final learned $\mathcal{S}_1$ satisfies the constraint $\mathcal{S}_1=\Phi(\mathbf{S},\cdots,\mathbf{S})$. This is an inherent limitation of the tensor learning process. However, we argue that this constraint does not conflict with the final learned result, as it represents an idealized prior assumption, whereas the learned result reflects a practical representation. Minor deviations from the ideal constraint are acceptable within the learning process. To support this, we have conducted heatmap experiments in the experimental section, which demonstrate that each frontal slice of the learned $\mathcal{S}_1$ exhibits only minor deviations, implying that $\mathcal{S}_1=\Phi(\mathbf{S},\cdots,\mathbf{S})$ approximately holds.   

%\textbf{\textit{Remark 4:}} It is worth noting that our proposed graph fusion strategy has a significant difference from the conventional consensus graph fusion method. In the traditional graph fusion method, a connectivity constraint is imposed on the objective optimization to ensure that the learned consensus graph comprises exactly $c$ connected components. However, in our method, we discard this constraint for simplicity. Instead, after obtaining the learned fusion graph $\mathbf{S}_a$, we construct the affinity matrix $\mathbf{S}_{af}$ using $\mathbf{S}_{af}=\frac{\mathbf{S}_a+\mathbf{S}^T_a}{2}$, and then perform spectral clustering.   

\begin{algorithm}[tb]
\caption{Procedures for optimization in CSTGL}
\label{alg: learning S}
\textbf{Input: }Multi-view data $\{\mathbf{X}^{(v)}\}^m_{v=1}$,
parameters $\alpha$, $\beta$ and $\gamma$.\\
\textbf{Initialize: } Initialize $\mathbf{A}^{(v)}$ with probabilistic $k$-nearest neighbors; $\mathbf{W}^{(v)}=\mathbf{0}$, $\mathbf{K}^{(v)}=\mathbf{0}$, $\mathbf{S}^{(v)}_1=\mathbf{0}$, $\mathbf{S}^{(v)}_2=\mathbf{0}$, $\mathbf{E}^{(v)}=\mathbf{0}$, $\mathbf{Q}^{(v)}_1=\mathbf{0}$, $\mathbf{Q}^{(v)}_2=\mathbf{0}$, $\mathbf{Q}^{(v)}_3=\mathbf{0}$; $\mu=2$, $\rho=0.1$; $\mathbf{S}_a = \mathbf{0}$.
\begin{algorithmic}[1] %[1] enables line numbers
\WHILE{not converged}
	\FOR{$\forall v=1,\cdots,m$}
		\STATE Update $\mathbf{A}^{(v)}$ according to \textit{subproblem 1}.
	\ENDFOR
\STATE Update $\mathcal{W}$, $\mathcal{K}$, $\mathcal{S}_2$, $\mathcal{S}_1$, $\mathcal{E}$ according to \textit{subproblems 2-6}.
\STATE Update multipiers $\mathcal{Q}_1$, $\mathcal{Q}_2$, $\mathcal{Q}_3$, $\rho$ according to \textit{subproblem 7}.
%\STATE Update $\rho$ by $\rho=\mu\rho$.
%\IF {$\Vert\mathbf{X}_a-\mathbf{PH}_a-\mathbf{E}_1\Vert_{\infty}<tol$ and $\Vert\mathbf{H}_a-\mathbf{H}_a\mathbf{Z}_a-\mathbf{E}_2\Vert_{\infty}<tol$ and $\Vert\mathbf{J}-\mathbf{Z}_a\Vert_{\infty}<tol$}
%\STATE Converged.
%\ENDIF
%\STATE $t\gets t+1$.
\ENDWHILE
\STATE Calculate the final graph matrix $\mathbf{S}_a$ using Eq. (\ref{eq036}).
\end{algorithmic}
\textbf{Output: }The learned fusion graph $\mathbf{S}_a$.
\end{algorithm}

\subsection{Computational Complexity}

The main complexity of CSTGL contains the cost of solving 6 subproblems. As the update of Lagrange multipliers involves matrix addition operations, we ignore its complexity. The detailed computational complexity analysis is presented below. For the $\mathbf{A}^{(v)}-$subproblem, the complexity of calculating (\ref{eq021}) for a single view takes $O(n^2)$, resulting in a total computational complexity of $O(mn^2)$ for all views. For the $\mathcal{W}-$subproblem, it consumes a complexity of $O(mn^2)$. For the $\mathcal{K}-$subproblem, the complexity includes performing the FFT and inverse FFT of a $n\times n\times m$ tensor, as well as executing the SVD in the Fourier domain, which requires $O(mn^3+mn^2\log(m))$ \cite{wu2019essential}. The updates of $\mathcal{S}_2$, $\mathcal{S}_1$ and $\mathcal{E}$ have similar solving procedures, all involving FFT and inverse FFT operations, which takes $O(mn^2\log(m))$. Therefore, the entire complexity of solving CSTGL is $O(mn^3+mn^2\log(m)+mn^2)$.        

\section{Experiments}

In this section, we conduct comprehensive experiments to validate the effectiveness of CSTGL. To be specific, we carry out the clustering performance and affinity matrix visualization comparison. We also conduct the parameter sensitivity and convergence analysis, and compare the running time of various methods. All experiments are conducted on a computer equipped with a 12th Generation Intel(R) Core(TM) i5-12490F CPU. Experimental results are obtained by averaging 10 trials. 

\begin{figure*}[htbp]
\centering
\includegraphics[width=0.85\textwidth]{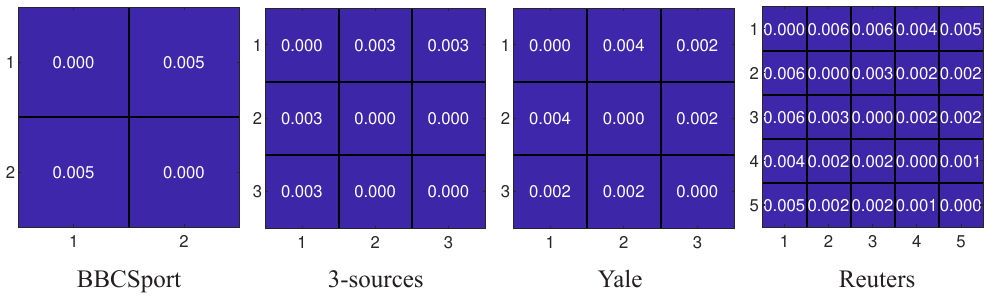} % Reduce the figure size so that it is slightly narrower than the column. Don't use precise values for figure width.This setup will avoid overfull boxes.
\caption{The Frobenius norm heatmaps between $\mathcal{S}_1^{(i)}$ and $\mathcal{S}_1^{(j)}$ on different datasets.}
\label{hotmap}
\end{figure*}  

%
%\begin{table}[!htb]
%    \centering
%    \small
%    \begin{tabular}{lrrrr}
%    \hline
%        Dataset & View & Sample & Dimension of features \\
%    \hline
%    	HW            & 2 & 2000  & {784, 256} \\
%    	Reuters       & 5 & 600   & {21526, 24892, 34121, 15487, 11539} \\
%        BBCSport      & 2 & 544   & {3183, 3203} \\
%        3-sources     & 3 & 169   & {3560, 3631, 3068} \\
%        Yale          & 3 & 165   & {4096, 3304, 6750} \\
%        WebKB         & 3 &	203   & {1703, 230, 230} \\
%    \hline
%    \end{tabular}
%     \caption{Summary of multi-view datasets}
%    \label{tab:dataset}
%\end{table}

\begin{table*}[h!]
    \centering
    \caption{Evaluation metrics of all compared methods on HW, Reuters and BBCSport ($\%$). Bold font indicates the highest performance, and underline font indicates the second-best performance.}
    \resizebox{\linewidth}{!}{
    \begin{tabular}{ c c c c c c c c c c c c c }
    \hline
        Datasets & \multicolumn{4}{c}{HW} & \multicolumn{4}{c}{Reuters} & \multicolumn{4}{c}{BBCSport} \\
    % \cline{2-5}{0.4pt}
    \cmidrule(lr){2-5} \cmidrule(lr){6-9} \cmidrule(lr){10-13}
        Method & ACC & NMI & ARI & Fscore & ACC & NMI & ARI & Fscore & ACC & NMI & ARI & Fscore \\
    \hline
        MCGC      & 0.6890 & 0.6684 & 0.5653 & 0.6116 & 0.3167 & 0.2510 & 0.0431 & 0.2894 & 0.9743 & 0.9145 & 0.9376 & 0.9525 \\
        CSMSC     & 0.8959 & 0.8252 & 0.7944 & 0.8115 & 0.5043 & 0.3271 & 0.2288 & 0.3660 & 0.9559 & 0.8619 & 0.8882 & 0.9148 \\
        LMSC      & 0.8326 & 0.8152 & 0.7344 & 0.7605 & 0.3789 & 0.2711 & 0.1345 & 0.3246 & 0.8957 & 0.7953 & 0.7722 & 0.8325 \\
        MCLES     & 0.8277 & 0.8898 & 0.8054 & 0.8256 & 0.3721 & 0.3344 & 0.1298 & 0.3476 & 0.8908 & 0.8351 & 0.8475 & 0.8933 \\
        t-SVD-MSC & 0.9885 & 0.9859 & 0.9867 & 0.9870 & 0.9450 & 0.8819 & 0.8751 & 0.8958 & 0.9706 & 0.9049 & 0.9172 & 0.9371 \\
        WTSNM     & 0.6565 & 0.5804 & 0.5040 & 0.5539 & 0.3717 & 0.1515 & 0.1099 & 0.2576 & 0.5074 & 0.2348 & 0.1672 & 0.3486 \\
        TBGL      & 0.9805 & 0.9604 & 0.9554 & 0.9598 & 0.2133 & 0.1265 & 0.0068 & 0.2822 & 0.5202 & 0.2508 & 0.1340 & 0.4276 \\
        LTBPL     & \underline{0.9990} & \underline{0.9973} & \underline{0.9978} & \underline{0.9980} & \underline{0.9600} & \underline{0.9180} & \underline{0.9086} & \underline{0.9238} & \textbf{0.9982} & \textbf{0.9924} & \textbf{0.9941} & \textbf{0.9955} \\
        UDBGL     & 0.6770 & 0.5900 & 0.4791 & 0.5449 & 0.3583 & 0.1412 & 0.1154 & 0.2908 & 0.4963 & 0.1940 & 0.2425 & 0.4865 \\
        MMGC      & 0.5910 & 0.5321 & 0.3999 & 0.4621 & 0.6367 & 0.4445 & 0.3860 & 0.4931 & 0.9596 & 0.8753 & 0.8931 & 0.9183 \\
        SLMVGC    & 0.7205 & 0.6880 & 0.5938 & 0.6363 & 0.3693 & 0.1348 & 0.0743 & 0.2554 & 0.7478 & 0.5763 & 0.4871 & 0.6060 \\
        \textbf{Ours} & \textbf{1.0000} & \textbf{1.0000} & \textbf{1.0000} & \textbf{1.0000} & \textbf{0.9967} & \textbf{0.9896} & \textbf{0.9920} & \textbf{0.9933} & \underline{0.9963} & \underline{0.9854} & \underline{0.9907} & \underline{0.9929} \\
    \hline
    \end{tabular}
    }
    \label{table:results1}
\end{table*}

\begin{table*}[h!]
    \centering
    \caption{Evaluation metrics of all compared methods on 3-sources, Yale and WebKB ($\%$). Bold font indicates the highest performance, and underline font indicates the second-best performance.}
    \resizebox{\linewidth}{!}{
    \begin{tabular}{ c c c c c c c c c c c c c }
    \hline
        Datasets & \multicolumn{4}{c}{3-sources} & \multicolumn{4}{c}{Yale} & \multicolumn{4}{c}{WebKB} \\
    \cmidrule(lr){2-5} \cmidrule(lr){6-9} \cmidrule(lr){10-13}
        Method & ACC & NMI & ARI & Fscore & ACC & NMI & ARI & Fscore & ACC & NMI & ARI & Fscore \\
    \hline
        MCGC      & 0.4793 & 0.2737 & 0.0879 & 0.3911 & 0.6242 & 0.6517 & 0.4409 & 0.4767 & 0.7241 & 0.3016 & 0.3767 & 0.6489 \\
        CSMSC     & 0.6509 & 0.4836 & 0.4658 & 0.5840 & 0.7044 & 0.7135 & 0.5421 & 0.5637 & 0.6502 & 0.2985 & 0.3313 & 0.6085 \\
        LMSC      & 0.5266 & 0.3897 & 0.3097 & 0.4603 & 0.7421 & 0.7538 & 0.5844 & 0.6101 & 0.7340 & 0.3107 & 0.3926 & 0.6592 \\
        MCLES     & 0.6864 & 0.5904 & 0.4461 & 0.6100 & 0.6976 & 0.7290 & 0.5005 & 0.5341 & 0.6976 & \textbf{0.7290} & \underline{0.5005} & 0.5341 \\
        t-SVD-MSC & 0.6391 & 0.6043 & 0.4958 & 0.6007 & \underline{0.7455} & \textbf{0.8011} & 0.6237 & 0.6481 & 0.7143 & 0.4132 & 0.4503 & 0.6650 \\
        WTSNM     & 0.4024 & 0.3522 & 0.1590 & 0.3407 & 0.7333 & 0.7186 & 0.5101 & 0.5416 & 0.6108 & 0.2546 & 0.2731 & 0.5509 \\
        TBGL      & 0.3491 & 0.1011 & 0.0143 & 0.3683 & 0.6909 & 0.7253 & 0.5006 & 0.5347 & 0.6305 & 0.2734 & 0.1919 & 0.5401 \\
        LTBPL     & 0.6686 & 0.4372 & 0.4134 & 0.5985 & \underline{0.7455} & 0.7868 & \underline{0.6358} & \underline{0.6589} & 0.5714 & 0.0666 & 0.0556 & 0.5755 \\
        UDBGL     & 0.4142 & 0.1720 & 0.1031 & 0.3410 & 0.5515 & 0.6227 & 0.3761 & 0.4192 & \underline{0.7685} & 0.3902 & 0.4817 & \underline{0.7072} \\
        MMGC      & \underline{0.7219} & \underline{0.5758} & \underline{0.5221} & \underline{0.6178} & \underline{0.7455} & 0.7557 & 0.5675 & 0.5957 & 0.6650 & 0.4070 & 0.4269 & 0.6084 \\
        SLMVGC    & 0.6219 & 0.4540 & 0.3650 & 0.5214 & 0.6224 & 0.6292 & 0.4042 & 0.4417 & 0.7094 & 0.2767 & 0.3194 & 0.6363 \\
        \textbf{Ours} & \textbf{0.7757} & \textbf{0.6797} & \textbf{0.6264} & \textbf{0.7202} & \textbf{0.7697} & \underline{0.7956} & \textbf{0.6381} & \textbf{0.6608} & \textbf{0.7733} & \underline{0.4213} & \textbf{0.5309} & \textbf{0.7335} \\
    \hline
    \end{tabular}
    }
    \label{table:results2}
\end{table*}

\subsection{Experimental Settings}

\subsubsection{Datasets} We select six representative real-world multi-view datasets, and a brief description on them is given below. 

%{\color{blue}{Additionally, we include a table summarizing the characteristics of these multi-view datasets in Table \ref{tab:dataset}}}.

\textbf{HW} \footnotemark \footnotetext{https://cs.nyu.edu/$\sim$ roweis/data.html}: It is composed of 2,000 data points for digits 0 and 9 from UCI machine learning repository. The digit images are obtained from two sources: MNIST and USPS Handwritten Digits. Accordingly, two views are constructed, with feature dimensions of 784 and 254, respectively.    

\textbf{Reuters} \footnotemark \footnotetext{https://lig-membres.imag.fr/grimal/data.html}: It contains 18,785 documents written in five different languages: French, Italian, English, German, and Spanish. Each view corresponds to a document representation in one of these languages. Following the widely employed selection strategy \cite{brbic2018multi}, we use 600 documents from 6 categories in our experiments.

\textbf{BBCSport} \footnotemark \footnotetext{http://mlg.ucd.ie/datasets/bbc.html}: This dataset contains 544 documents sourced from the BBC Sport website, covering news from five different categories. Each view represents a different feature representation of the documents. In the experiments, two views with dimensions of 3,183 and 3,203 are selected.

\textbf{3-sources} \footnotemark \footnotetext{http://mlg.ucd.ie/datasets/3sources.html}: This dataset includes 948 news articles covering 416 unique news stories, collected from three well-known online news sources. Each view corresponds to articles from one of these sources. Following the existing literature \cite{brbic2018multi, jiang2022tensorial}, we use 169 articles that are classified into 6 categories across all three sources. 

\textbf{Yale} \footnotemark \footnotetext{http://vision.ucsd.edu/content/yale-face-database}: It consists of 165 gray-scale images of 15 individuals with different facial expressions and configurations. Followed by \cite{chen2022low}, three views are constructed based on different feature representations: a 4,096-dimensional intensity feature, a 3,304-dimensional LBP feature, and a 6,750-dimensional Gabor feature.

\textbf{WebKB} \footnotemark \footnotetext{http://www.webkb.org/webkb.html}: This dataset consists of 203 web-pages of 4 classes. Each view represents a different textual aspect of a web page, including its main content, the anchor text of hyperlinks, and the text in its title.

\subsubsection{Compared Methods} Several multi-view baselines are involved to perform comparison, they are: 

%\textbf{MCGC} \cite{zhan2018multiview}, \textbf{CSMSC} \cite{luo2018consistent}, \textbf{LMSC} \cite{zhang2018generalized}, \textbf{MCLES} \cite{chen2020multi}, \textbf{t-SVD-MSC} \cite{xie2018unifying}, \textbf{WTSNM} \cite{xia2021multiview}, \textbf{TBGL} \cite{xia2022tensorized}, \textbf{LTBPL} \cite{chen2022low}, \textbf{UDBGL} \cite{fang2023efficient}, \textbf{MMGC} \cite{tan2023metric}, and \textbf{SLMVGC} \cite{tan2023sample}.         

\textbf{MCGC} \cite{zhan2018multiview}: This method proposes an effective disagreement cost function to learn the consensus graph.    

\textbf{CSMSC} \cite{luo2018consistent}: This method takes into account both consistent and specific information to perform subspace clustering.

\textbf{LMSC} \cite{zhang2018generalized}: This method aims to learn a latent representation for comprehensive information recovery. 

\textbf{MCLES} \cite{chen2020multi}: This method clusters multi-view data in a learned latent embedding space.

\textbf{t-SVD-MSC} \cite{xie2018unifying}: This method applies the t-SVD based tensor nuclear norm to capture high-order information among views.    

\textbf{WTSNM} \cite{xia2021multiview}: This method investigates the weighted tensor Schatten $p$-norm in the tensor-singular value decomposition.  

\textbf{TBGL} \cite{xia2022tensorized}: This method designs a variance-based de-correlation anchor selection strategy for bipartite graph.

\textbf{LTBPL} \cite{chen2022low}: This method integrates the low-rank probability matrices and the consensus indicator graph into a framework.

\textbf{UDBGL} \cite{fang2023efficient}: This method unifies the learning of the view-specific and consensus bipartite graph.

\textbf{MMGC} \cite{tan2023metric}: This method integrates the metric learning and graph learning to simplify the relationship among data.

\textbf{SLMVGC} \cite{tan2023sample}: This method attempts to explore the topological structure of data for performance improvement.  

\begin{figure*}[htbp]
\centering
\includegraphics[width=0.82\textwidth]{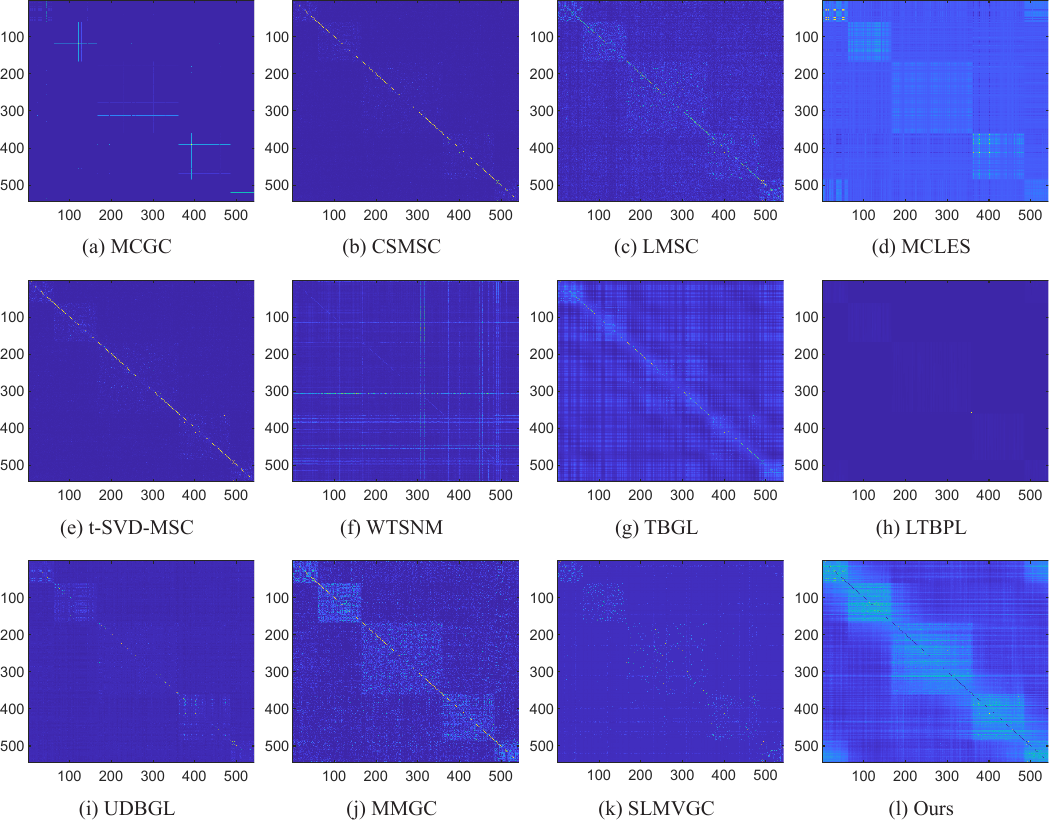} % Reduce the figure size so that it is slightly narrower than the column. Don't use precise values for figure width.This setup will avoid overfull boxes.
\caption{Visualization of affinity matrix on BBCSport.}
\label{Aff_matrix}
\end{figure*} 

\begin{figure}[htbp]
\centering
\includegraphics[width=0.5\textwidth]{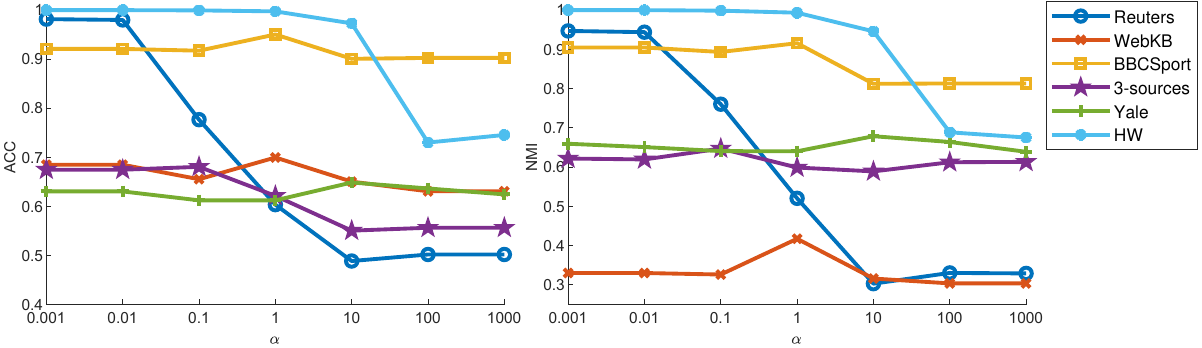} % Reduce the figure size so that it is slightly narrower than the column. Don't use precise values for figure width.This setup will avoid overfull boxes.
\caption{Sensitivity analysis of $\alpha$ on different datasets, and $\beta=100$, $\gamma=10$.}
\label{para_alpha}
\end{figure}

\subsubsection{Evaluation Metrics and Parameter Setting} All the methods are evaluated by : clustering accuracy (ACC), normalized mutual information (NMI), adjusted rand index (ARI), and F1 score (Fscore). For a fair comparison, we carefully tune the parameters from the grid set $\{10^{-3},10^{-2},10^{-1},10^0,10^1,10^2,10^3\}$ to achieve the optimal performance for all methods. Alternatively, we utilize recommended parameters if they are available for previous methods.

\subsection{Experimental Results}

\subsubsection{Verification of Consistent Tensor Constraint} To demonstrate that each frontal slice of the learned $\mathcal{S}_1$ exhibits only minor deviations, we compare the Frobenius norm heatmaps of any two learned slices, $\mathbf{S}_1^{(i)}$ and $\mathbf{S}_1^{(j)}$, as shown in Fig. \ref{hotmap}. It is clearly seen that on the tested datasets, every non-diagonal block displays very small values, indicating that the differences between any two learned slices are negligible. This empirical analysis confirms that the frontal matrices of the learned consistent tensor graph $\mathcal{S}_1$ are nearly identical, i.e., $\mathcal{S}_1=\Phi(\mathbf{S},\cdots, \mathbf{S})$ approximately holds.

\begin{table}[h!]
    \centering
    \caption{Ablation studies of CSTGL using different distance measures on various datasets, and the metric ACC is employed. (a) Euclidean distance, and (b) pseudo-Stiefel manifold distance.}
    \begin{tabular}{ccccccc}
    \hline
     dis  & HW   & Reuters & BBCSport & 3-sources & Yale & Webkb \\
    \hline
    (a) & \textbf{1.0000} & 0.9933 & 0.9467 & \textbf{0.7870} & 0.7479 & 0.7291 \\
    (b) & \textbf{1.0000} & \textbf{0.9983} & \textbf{0.9963} & 0.7757 & \textbf{0.7697} & \textbf{0.7783} \\
    \hline
    \end{tabular}
    \label{tab:distance}
\end{table}

\begin{table*}[h!]
    \centering
    \caption{Ablation studies of tensor graph fusion on the BBCSport, 3-sources and WebKB datasets. (a) retaining consistent tensor graph, (b) retaining view-specific tensor graph, and (c) retaining both consistent and view-specific tensor graphs.}
    \resizebox{\linewidth}{!}{
    \begin{tabular}{c|cc|cccc|cccc|cccc}
    \hline
        \multirow{2}{2em}{Cases} & \multicolumn{2}{c|}{Components} & \multicolumn{4}{c|}{BBCSport} & \multicolumn{4}{c|}{3-sources} & \multicolumn{4}{c}{WebKB} \\
        & Con. & Spe. & ACC & NMI & ARI & Fscore & ACC & NMI & ARI & Fscore & ACC & NMI & ARI & Fscore \\
    \hline
        (a) & \ding{51} & & 0.7188 & 0.8709 & 0.7425 & 0.7987 & 0.6094 & 0.6401 & 0.4367 & 0.5491 & 0.7290 & \textbf{0.4879} & 0.5293 & 0.7277 \\
        (b) & & \ding{51} & 0.3217 & 0.0844 & 0.0474 & 0.2545 & 0.6745 & 0.6267 & 0.4673 & 0.5725 & 0.0321 & 0.0460 & 0.0053 & 0.3141 \\
        (c) & \ding{51} & \ding{51} & \textbf{0.9963} & \textbf{0.9854} & \textbf{0.9907} & \textbf{0.9929} & \textbf{0.7751} & \textbf{0.6783} & \textbf{0.6252} & \textbf{0.7193} & \textbf{0.7733} & 0.4213 & \textbf{0.5309} & \textbf{0.7335} \\
    \hline
    \end{tabular}
    }
    \label{tab:consistency and specifity}
\end{table*}

\begin{table*}[h!]
    \centering
    \caption{Ablation studies of tensor graph fusion on the Yale, Reuters and HW datasets. (a) retaining consistent tensor graph, (b) retaining view-specific tensor graph, and (c) retaining both consistent and view-specific tensor graphs.}
    \resizebox{\linewidth}{!}{
    \begin{tabular}{c|cc|cccc|cccc|cccc}
    \hline
        \multirow{2}{2em}{Case} & \multicolumn{2}{c|}{Component} & \multicolumn{4}{c|}{Yale} & \multicolumn{4}{c|}{Reuters} & \multicolumn{4}{c}{HW} \\
        & Con. & Spe. & ACC & NMI & ARI & Fscore & ACC & NMI & ARI & Fscore & ACC & NMI & ARI & Fscore \\
    \hline
    % 0.642424242424243	0.726861193649420	0.533546373145148	0.563811684628474
    % 0.339393939393939	0.395674817357482	0.109788780377016	0.166279069767442
    % 
        (a) & \ding{51} & & 0.6545 & 0.7392 & 0.5530 & 0.5819 & 0.9950 & 0.9844 & 0.9880 & 0.9900 & 0.9995 & 0.9986 & 0.9989 & 0.9990 \\
        (b) & & \ding{51} & 0.3273 & 0.3766 & 0.0990 & 0.1567 & 0.2133 & 0.0135 & 0.0014 & 0.1668 & 0.2110 & 0.0742 & 0.0353 & 0.1322 \\
        (c) & \ding{51} & \ding{51} & \textbf{0.7697} & \textbf{0.7956} & \textbf{0.6381} & \textbf{0.6608} & \textbf{0.9967} & \textbf{0.9896} & \textbf{0.9920} & \textbf{0.9933} & \textbf{1.0000} & \textbf{1.0000} & \textbf{1.0000} & \textbf{1.0000} \\
    \hline
    \end{tabular}
    }
    \label{tab:consistency and specifity 2}
\end{table*}  

\subsubsection{Clustering Comparison} We record the evaluation metrics of different methods on six datasets in Tables \ref{table:results1} and \ref{table:results2}. In particular, Table \ref{table:results1} reports the clustering results on the HW, Reuters and BBCsport datasets, while Table \ref{table:results2} shows the clustering performance on the 3-sources, Yale and WebKB datasets. From the table results, we have some findings below:
 
a) Our proposed CSTGL exhibits the best clustering performance on most tested datasets. Specifically, it achieves the top performance in terms of all the evaluation metrics on the HW, Reuters and 3-sources datasets. Simultaneously, it shows the best performance in terms of ACC, ARI and Fscore on the Yale and WebKB datasets. For example, CSTGL achieves a perfect score of $100\%$ in ACC, NMI, ARI and Fscore on the HW dataset. On the 3-sources dataset, CSTGL shows an accuracy of $77.57\%$, marking an improvement of $5.38\%$ compared to the second-performing method, MMGC. On Yale and WebKB, CSTGL continues to achieve the highest accuracy, reaching $76.97\%$ and $77.33\%$ respectively. Despite that CSTGL ranks the second-best method on the BBCsport, all of its evaluation metrics are quite close to those of the top-performing method, LTBPL.

b) Multi-view methods with both consistent and view-specific information considered generally outperform those methods that only consider consistency among views. For example, CSMSC, a representative subspace clustering method that exploits both consistency and specificity, behaves better than MCGC on the HW, Reuters, 3-sources, and Yale datasets. It achieves an accuracy improvement of $20.69\%$, $18.76\%$, $17.16\%$, and $8.02\%$, respectively. In addition, methods based on latent representation, such as LMSC and MCLES, while not offering the best, still provide commendable performance. This originates from the fact that latent representation-based methods explore comprehensive information from the potential latent space. 

c) Generally, tensor-based multi-view methods outperform other non-tensor-based competing methods. For example, previous methods such as t-SVD-MSC and LTBPL demonstrate clustering performance that is only second to our proposed method on the HW dataset, achieving accuracies of $98.85\%$ and $99.9\%$, respectively. This superior performance can be attributed to the comprehensive consideration of high-order information by tensor decomposition. In contrast to these previous tensor-based methods, our proposed method, CSTGL, exhibits even better clustering performance. This can be largely attributed to its capability to capture the underlying structure of high-dimensional data, and the utilization of a newly designed tensor-based graph learning model that ensures both consistency and specificity, while effectively excluding the influence of noise.

In the supplementary material, we present additional clustering comparison experiments with synthetic noise added to multi-view data. The results further demonstrate our model's robustness against noise.

\begin{figure*}[htbp]
\centering
\includegraphics[width=0.85\textwidth]{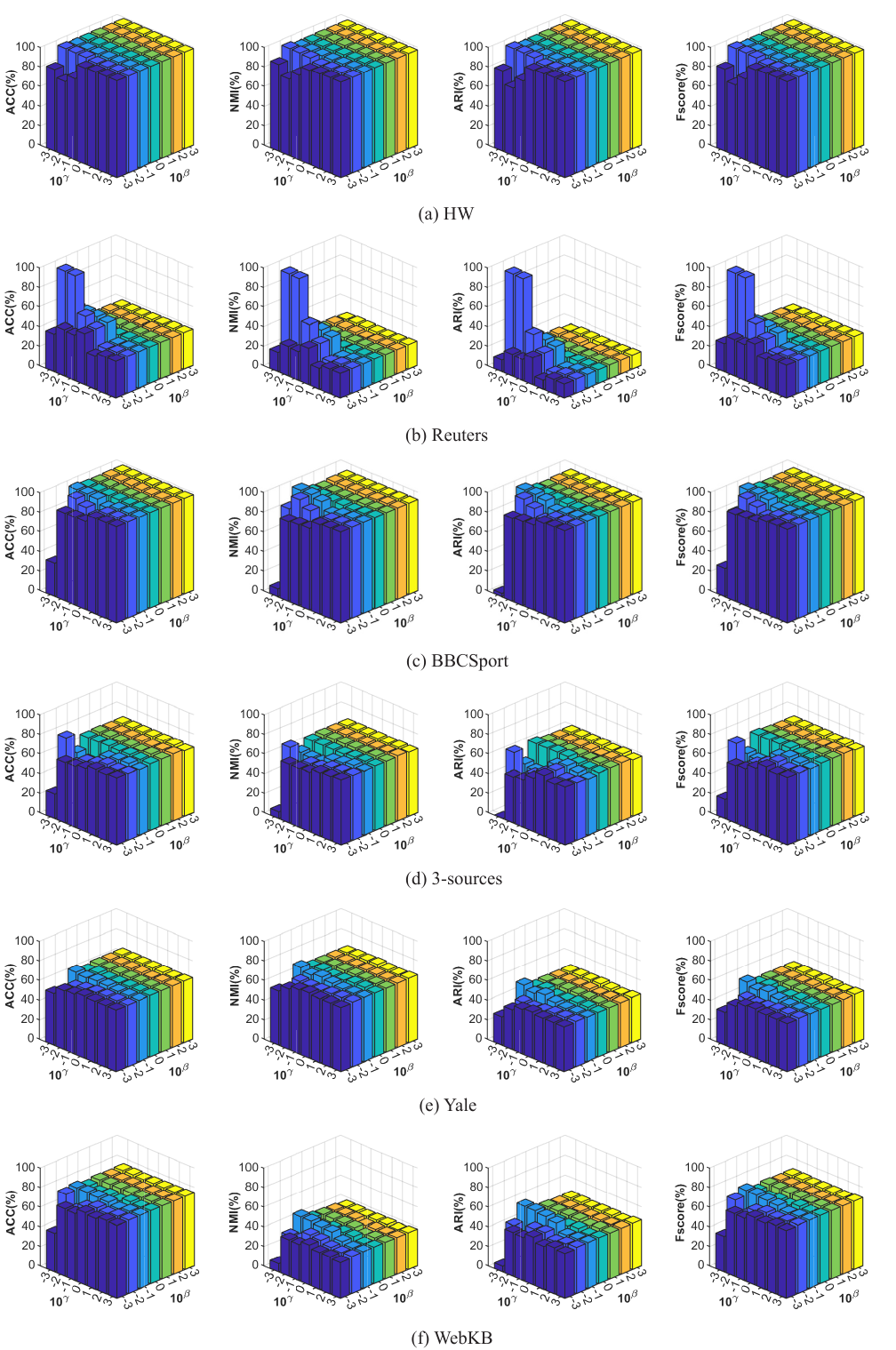} % Reduce the figure size so that it is slightly narrower than the column. Don't use precise values for figure width.This setup will avoid overfull boxes.
\caption{Parameter sensitivity results on six datasets, and $\alpha=10$.}
\label{para_sen}
\end{figure*}

\begin{figure*}[htbp]
\centering
\includegraphics[width=0.85\textwidth]{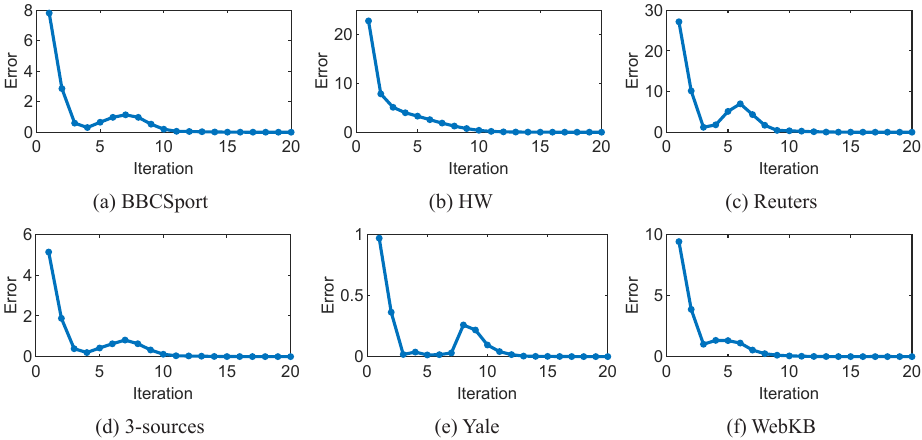} % Reduce the figure size so that it is slightly narrower than the column. Don't use precise values for figure width.This setup will avoid overfull boxes.
\caption{Convergent behavior of CSTGL.}
\label{convergence}
\end{figure*} 

\begin{figure*}[htbp]
\centering
\includegraphics[width=0.85\textwidth]{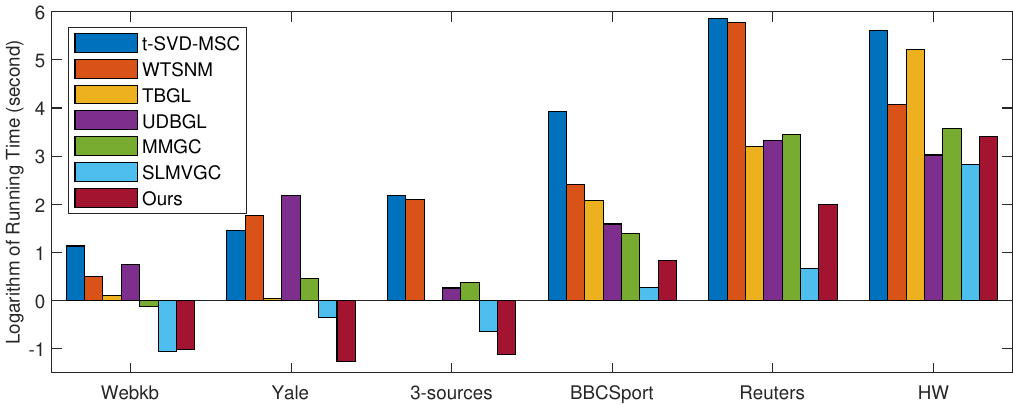} % Reduce the figure size so that it is slightly narrower than the column. Don't use precise values for figure width.This setup will avoid overfull boxes.
\caption{Comparison of running time of various multi-view methods.}
\label{running_time}
\end{figure*} 

\subsubsection{Ablation Studies}

To validate the advantages of similarity measure using pseudo-Stiefel manifold distance and tensor graph fusion, we conduct corresponding ablation studies. Specifically, when investigating the effects of using different distance measures, we alternate the similarity measure between Euclidean distance and pseudo-Stiefel manifold distance while maintaining all other implementation procedures unchanged. When exploring the benefits of tensor graph fusion, we examine the effects of discarding the consistent tensor $\mathcal{S}_1$ and specific tensor $\mathcal{S}_2$, individually. 

Table \ref{tab:distance} records the experimental results of CSTGL using different distance measures on various datasets. It is seen that for most tested datasets, the performance of CSTGL using Euclidean distance is inferior to that using pseudo-Stiefel manifold distance. However, for the 3-sources dataset, CSTGL exhibits weaker performance when using the pseudo-Stiefel manifold distance. This can be attributed to the nature of the 3-sources dataset, which is news-related, and the data may be sparse due to the limitation of news sources. In such a scenario, Euclidean distance, without considering the intrinsic structure of the data, could be more effective.  

Tables \ref{tab:consistency and specifity} and \ref{tab:consistency and specifity 2} shows the ablation studies of tensor graph fusion on six datasets. From an overall perspective, both the consistent tensor graph and specific tensor graph contribute to the performance improvement of CSTGL. In particular, on the BBCSport, WebKB, Yale, Reuters, and HW datasets, the consistent component contributes more significantly than the specific component, while on the 3-sources dataset, the specific component has a slightly greater contribution than the consistent component.

\subsubsection{Visualization of Affinity Matrix}

In Fig. \ref{Aff_matrix}, we exhibit the visualization results of affinity matrices for all the compared multi-view methods on BBCsport. In the graph-based methods, these affinity matrices are commonly known similarity graphs. It is seen that by comparing the block-diagonal structures of various multi-view methods, our method exhibits a more distinct block-diagonal structure, despite having some shadows in the off-diagonal blocks. This further demonstrates that CSTGL can achieve satisfactory clustering performance.  
     
\subsubsection{Parameter Sensitivity Analysis} To measure the influence of parameter choices on the CSTGL's performance, the parameter sensitivity analyses involving all the evaluation metrics on six datasets are conducted. We first study the influence of $\alpha$ by fixing $\beta$ and $\gamma$. As shown in Fig. \ref{para_alpha}, the performance of CSTGL remains insensitive to $\alpha$ on the WebKB, BBCSport, 3-sources, and Yale datasets. However, a noticeable decline in performance is observed with large $\alpha$ on the Reuters and HW datasets. These observations indicate flexibility in choosing $\alpha$, with a preference of selecting small $\alpha$ for some real-world datasets.

We then investigate the sensitivity induced by $\beta$ and $\gamma$, and the experimental results are shown in Fig. \ref{para_sen}. Overall, it is observed that the performance of CSTGL does not significantly vary with the choice of $\beta$. However, it does exhibit sensitivity to the choice of $\gamma$, particularly when $\beta$ is set to a small value. More specifically, when $\beta=10^{-3}$, a decrease in $\gamma$ leads to a deterioration in CSTGL's performance. By referring to the objective function in (\ref{eq014}), we can see that the parameter $\gamma$ is used to regulate the view-specific tensor term. Therefore, we recommend using large $\gamma$ to emphasize the view's specificity for desirable clustering performance.   

\subsubsection{Empirical Convergence Analysis} In multi-view clustering, ensuring consistency is an essential prerequisite. Therefore, focusing on the update related to consistency is a straightforward approach when investigating convergence. As a result, we plot the evolutionary curves of $\lVert\mathcal{S}_1^{k+1}-\mathcal{S}_1^k\rVert^2_F$ on six datasets. From Fig. \ref{convergence}, we see that the update of the consistent tensor $\mathcal{S}_1$ is convergent, arriving at the steady-state within $20$ iterations. Although $\mathcal{S}_1$ experiences an ascent within the iteration range of $[5,10]$ on some datasets, it generally converges to the steady-state after $10$ iterations. Slight oscillations during the convergence process of optimization are expected, especially when using gradient descent in the presence of noisy data.

\subsubsection{Running Time} To intuitively show the computational cost of CSTGL, we plot a bar chart to compare the actual running times of various multi-view methods, as shown in Fig. \ref{running_time}. The Y-axis of the bar chart represents the logarithmic running time. Our method nearly achieves the fastest running time on the WebKB, Yale, and 3-sources datasets. The logarithmic running time is around $-1$,  which corresponds to an actual running time of approximate $0.35$ seconds. On the BBCsport, Reuters and HW datasets, our method continues to demonstrate a competitive advantage in terms of running time. More specifically, our CSTGL outperforms previous t-SVD-MSC and WTSNM which are two typical tensor-based methods. 

\section{Conclusion}

We propose a novel tensor-based graph learning framework, namely CSTGL. Our work makes two significant contributions in 1) the learning of neighbor graph, and 2) the model of graph fusion. Firstly, we learn the neighbor graph of each view based on using the pseudo-Stiefel manifold distance, which is empirically verified to more effectively recover the intrinsic structure than using Euclidean distance. Secondly, after obtaining the learned neighbor graphs for all views, we formulate a tensor-based graph fusion model by assuming that the neighbor graph of each view is composed of a consistent graph, a view-specific graph, and a noise term. This model considers both consistency and specificity, while avoiding the issue of potential noise contamination in the learned useful information. Additionally, thanks to the t-SVD based tensor nuclear norm, CSTGL is capable of capturing high-order correlations among different views. Extensive experiments have demonstrated that CSTGL outperforms some SOTA multi-view methods.

%\section*{Acknowledgment}

\ifCLASSOPTIONcaptionsoff
  \newpage
\fi

\bibliographystyle{IEEEtran}
\bibliography{IEEEabrv,IEEEexample}

\begin{IEEEbiography}[{\includegraphics[width=1in,height=1.25in,clip,keepaspectratio]{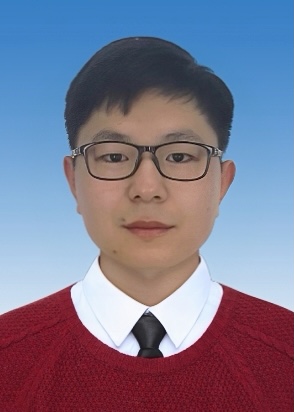}}]{Long Shi} (Member, IEEE) was born in Jiangsu Province, China. He received the Ph.D degree in electrical engineering from Southwest Jiaotong University, Chengdu, China, 2020. From 2018 to 2019, he was a visiting student with the Department of Electronic Engineering, University of York, U.K. Since 2021, he has been with the School of Computing and Artificial Intelligence, Southwestern University of Finance and Economics, where he is currently an associate professor. He has published several high-quality papers in IEEE TSP, IEEE TCSVT, IEEE SPL, IEEE TCSII, etc. His research interest lies at the intersection of signal processing and machine learning, including adaptive signal processing, multi-view (multimodal) learning, large language model.\end{IEEEbiography}

\begin{IEEEbiography}[{\includegraphics[width=1in,height=1.25in,clip,keepaspectratio]{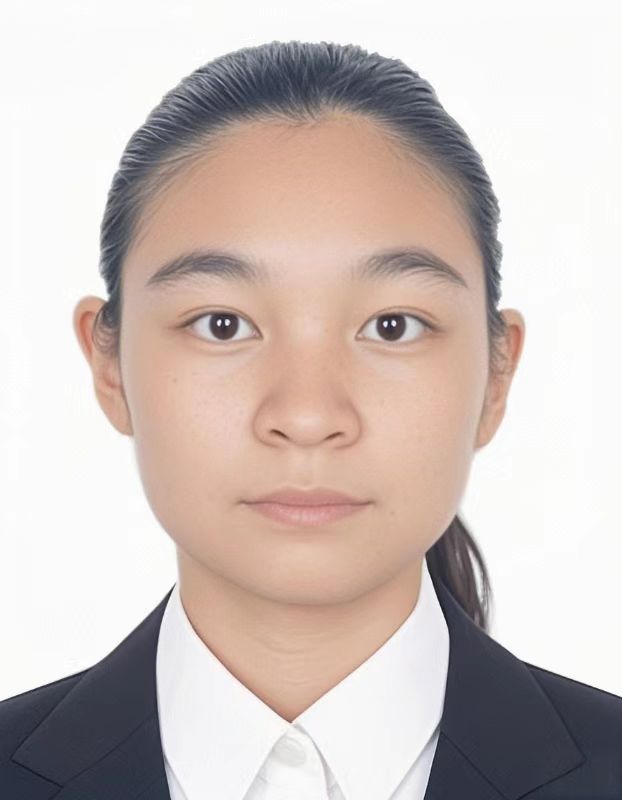}}]{Yunshan Ye} received the B.S. degree in 2021 from Capital University of Economics and Business, Beijing, China. She is currently working toward the Ph.D. degree with the School of Business Administration, Southwestern University of Finance and Economics, Chengdu, China. Her research interests include multi-view learning and deep learning.\end{IEEEbiography}

\begin{IEEEbiography}[{\includegraphics[width=1in,height=1.25in,clip,keepaspectratio]{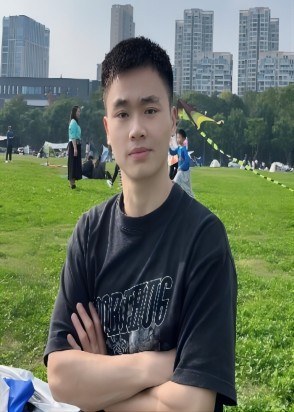}}]{Lei Cao} is currently pursuing a Master’s degree at School of Computing and Artificial Intelligence, Financial Intelligence and Financial Engineering Key Laboratory of Sichuan Province (FIFE), Southwestern University of Finance and Economics(SWUFE), supervised by Dr. Long Shi. His research interests are mainly in multi-view clustering, multimodal learning and applications in finance.\end{IEEEbiography}

\begin{IEEEbiography}[{\includegraphics[width=1in,height=1.25in,clip,keepaspectratio]{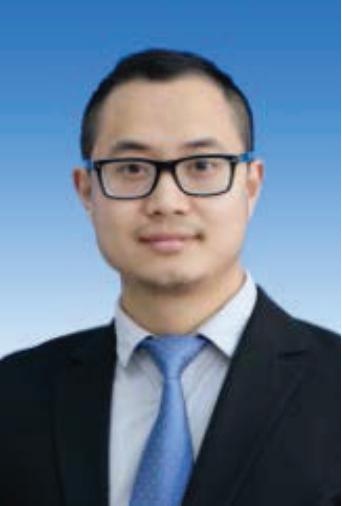}}]{Yu Zhao} received the B.S. degree from Southwest Jiaotong University in 2006, and the M.S. and Ph.D. degrees from the Beijing University of Posts and Telecommunications in 2011 and 2017, respectively. He is currently an Associate Professor at Southwestern University of Finance and Economics. His current research interests include machine learning, natural language processing, knowledge graph, Fintech. He has authored more than 30 papers in top journals and conferences including IEEE TKDE, IEEE TNNLS, IEEE TMC, ACL, ICME, etc.\end{IEEEbiography}

\begin{IEEEbiography}[{\includegraphics[width=1in,height=1.25in,clip,keepaspectratio]{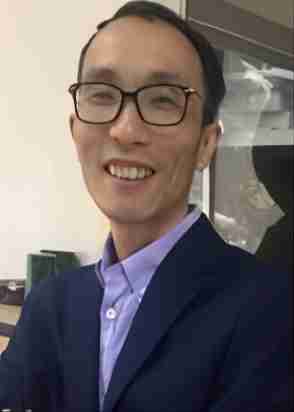}}]{Badong Chen} received the Ph.D. degree in Computer Science and Technology from Tsinghua University, Beijing, China, in 2008. He is currently a professor with the Institute of Artificial Intelligence and Robotics, Xi'an Jiaotong University, Xi'an, China. His research interests are in signal processing, machine learning, artificial intelligence and robotics. He has authored or coauthored over 300 articles in various journals and conference proceedings (with 13000+ citations in Google Scholar), and has won the 2022 Outstanding Paper Award of IEEE Transactions on Cognitive and Developmental Systems. Dr. Chen serves as a Member of the Machine Learning for Signal Processing Technical Committee of the IEEE Signal Processing Society, and serves (or has served) as an Associate Editor for several journals including IEEE Transactions on Neural Networks and Learning Systems, IEEE Transactions on Cognitive and Developmental Systems, IEEE Transactions on Circuits and Systems for Video Technology, Neural Networks and Journal of The Franklin Institute. He has served as a PC or SPC Member for prestigious conferences including UAI, IJCAI and AAAI, and served as a General Co-Chair of 2022 IEEE International Workshop on Machine Learning for Signal Processing.\end{IEEEbiography}

\end{document}